\def\year{2021}\relax
\documentclass[letterpaper]{article} % DO NOT CHANGE THIS
\usepackage{aaai21}  % DO NOT CHANGE THIS
\usepackage{times}  % DO NOT CHANGE THIS
\usepackage{helvet} % DO NOT CHANGE THIS
\usepackage{courier}  % DO NOT CHANGE THIS
\usepackage[hyphens]{url}  % DO NOT CHANGE THIS
\usepackage{graphicx} % DO NOT CHANGE THIS
\usepackage{natbib}  % DO NOT CHANGE THIS AND DO NOT ADD ANY OPTIONS TO IT
\usepackage{caption} % DO NOT CHANGE THIS AND DO NOT ADD ANY OPTIONS TO IT
\usepackage{xcolor}
\usepackage{amssymb}
\usepackage{times}
\usepackage{graphicx} % more modern
\usepackage{subcaption}
\usepackage{natbib}
\usepackage[lined,boxed,commentsnumbered]{algorithm2e}
\usepackage{mathtools}
\usepackage{paralist}
\usepackage{multirow}
\usepackage{float}

\newcommand{\R}{\mathbb{R}}
\newcommand{\T}{\mathcal{T}}

\newtheorem{theorem}{Theorem}

\newtheorem{proposition}[theorem]{Proposition}

\newcommand{\I}{\mathcal{I}}

\title{Sequential Attacks on Kalman Filter-based Forward Collision Warning Systems}
\author{
     Yuzhe Ma, Jon Sharp, Ruizhe Wang, Earlence Fernandes, Xiaojin Zhu\\
}
\affiliations{
Department of Computer Sciences, University of Wisconsin--Madison\\
$\{$yzm234, sharp-jr, ruizhe, earlence, jerryzhu$\}$@cs.wisc.edu\\
}

\begin{document}

\maketitle

% we enforce the edits to the measurement vectors:
% camera distance measurement plus attacker edit should be [0, 80]
% this is because the sensors we have pick up objects at distances upto 80 m but not beyond

% camera velocity measurement plus attacker edit should be [-30, 30] corresponding to 60 mph ego and stationary MIO
% this is physically plausible for example, freeway speeds.

\begin{abstract}
Kalman Filter (KF) is widely used in various domains to perform sequential learning or variable estimation. In the context of autonomous vehicles, KF constitutes the core component of many Advanced Driver Assistance Systems (ADAS), such as Forward Collision Warning (FCW). It tracks the states (distance, velocity etc.) of relevant traffic objects based on sensor measurements. The tracking output of KF is often fed into downstream logic to produce alerts, which will then be used by human drivers to make driving decisions in near-collision scenarios. In this paper, we study adversarial attacks on KF as part of the more complex machine-human hybrid system of Forward Collision Warning. Our attack goal is to negatively affect human braking decisions by causing KF to output incorrect state estimations that lead to false or delayed alerts. We accomplish this by sequentially manipulating measurements fed into the KF, and propose a novel Model Predictive Control (MPC) approach to compute the optimal manipulation. Via experiments conducted in a simulated driving environment, we show that the attacker is able to successfully change FCW alert signals through planned manipulation over measurements prior to the desired target time. These results demonstrate that our attack can stealthily mislead a distracted human driver and cause vehicle collisions.
\end{abstract}

% we enforce the edits to the measurement vectors:
% camera distance measurement plus attacker edit should be [0, 80]
% this is because the sensors we have pick up objects at distances upto 80 m but not beyond

% camera velocity measurement plus attacker edit should be [-30, 30] corresponding to 60 mph ego and stationary MIO
% this is physically plausible for example, freeway speeds.

%\begin{itemize}
%\item Introduce Kalman Filter and describe how is KF used in modern object tracking systems.

%\item Discuss the security concern of KF-based object tracking such as one-shot attack.

%\item Emphasize our main contribution of providing a general MPC-based framework for optimal attacks against KF. Highlight the two approaches that we solve optimal attack.
%\end{itemize}

\section{Introduction}
% ADAS is widely deployed and reduces road accidents. different sensors plus the strides in ML are used, for ex on camera. AI/ML techniques suffer from security issues. our goal is to understand what is the impact of those security issues on CPSs that use AI techniques.
Advanced Driver Assistance Systems (ADAS) are hybrid human-machine systems that are widely deployed on production passenger vehicles~\cite{nhtsa-adas}. They use sensing, traditional signal processing and machine learning to detect and raise alerts about unsafe road situations and rely on the human driver to take corrective actions. Popular ADAS examples include Forward Collision Warning (FCW), Adaptive Cruise Control and Autonomous Emergency Braking (AEB).  

Although ADAS hybrid systems are designed to increase road safety when drivers are distracted, attackers can negate their benefits by strategically tampering with their behavior. For example, an attacker could convince an FCW or AEB system that there is no imminent collision until it is too late for a human driver to avoid the crash. 

We study the robustness of ADAS to attacks. The core of ADAS typically involves tracking the states (e.g., distance and velocity) of road objects using Kalman filter (KF). Downstream logic uses this tracking output to detect unsafe situations before they happen. We focus our efforts on Forward Collision Warning (FCW), a popular ADAS deployed on production vehicles today. FCW uses KF state predictions to detect whether the ego vehicle (vehicle employing the ADAS system) is about to collide with the most important object in front of it and will alert the human driver in a timely manner. Thus, our concrete attack goal is to trick the KF that FCW uses and make it output incorrect state predictions that would induce false or delayed alerts depending on the specific physical situation. 

Recent work has examined the robustness of road object state tracking for autonomous vehicles~\cite{iclr-mot}. Their attacks create an instantaneous manipulation to the Kalman filter inputs without considering its sequential nature, the downstream logic that depends on filter output, or the physical dynamics of involved vehicles. This leads to temporarily hijacked Kalman filter state predictions that are incapable of ensuring that downstream logic is reliably tricked into producing false alerts. By contrast, we adopt an online planning view of attacking KFs that accounts for: (1) their sequential nature where current predictions depend on past measurements; and (2) the downstream logic that uses KF output to produce warnings. Our attack technique also considers a simplified model of human reaction to manipulated FCW warning lights. 

We propose a novel Model Predictive Control (MPC)-based attack that can sequentially manipulate measurement inputs to a KF with the goal of stealthily hijacking its behavior. Our attacks force FCW alerts that mask the true nature of the physical situation involving the vehicles until it is too late for a distracted human driver to take corrective actions. 

We evaluate our attack framework by creating a high-fidelity driving simulation using CARLA~\cite{carla}, a popular tool for autonomous vehicle research and development. We create test scenarios based on real-world driving data~\cite{nhtsa-guidelines,euro-ncap-protocol} and demonstrate the practicality of the attack in causing crashes involving the victim vehicle. Anonymized CARLA simulation videos of our attacks are available at \url{https://sites.google.com/view/attack-kalman-filter}.

% contributions
\noindent\textbf{Main Contributions:}
\begin{itemize}
\item We develop an optimal control-based attack against the popular FCW driver assistance system. Our attack targets several critical parts of the FCW pipeline -- Kalman filter tracking and prediction, FCW alert logic and human decision making in crash and near-crash scenarios.

\item We evaluate our control-based attacks in a high-fidelity simulation environment demonstrating that an attacker can compromise \emph{only} the camera-based measurement data and accomplish their goals of creating end-to-end unsafe situations for an FCW system, even under the constraint of limited manipulation to measurements.

\item We show that attack planning in advance of the targeted point is beneficial to the attack compared to without planning. Given 25 steps of planning (or 1.25 seconds based on specific physical situations in our evaluation) before the targeted time point, the attacker can cause the desired effect, while the attack fails without planning. Furthermore, via comparisons against a baseline greedy attack, we show that our attack can find near-optimal planning that achieves better overall performance. 

\end{itemize}

% longer-term vision: extension to pixel perturbation, partial attacker knowledge of model, physical.

% we enforce the edits to the measurement vectors:
% camera distance measurement plus attacker edit should be [0, 80]
% this is because the sensors we have pick up objects at distances upto 80 m but not beyond

% camera velocity measurement plus attacker edit should be [-30, 30] corresponding to 60 mph ego and stationary MIO
% this is physically plausible for example, freeway speeds.

\section{Background}
\begin{figure}[t!]
  \centering \includegraphics[width=1\columnwidth,height=0.443\textwidth]{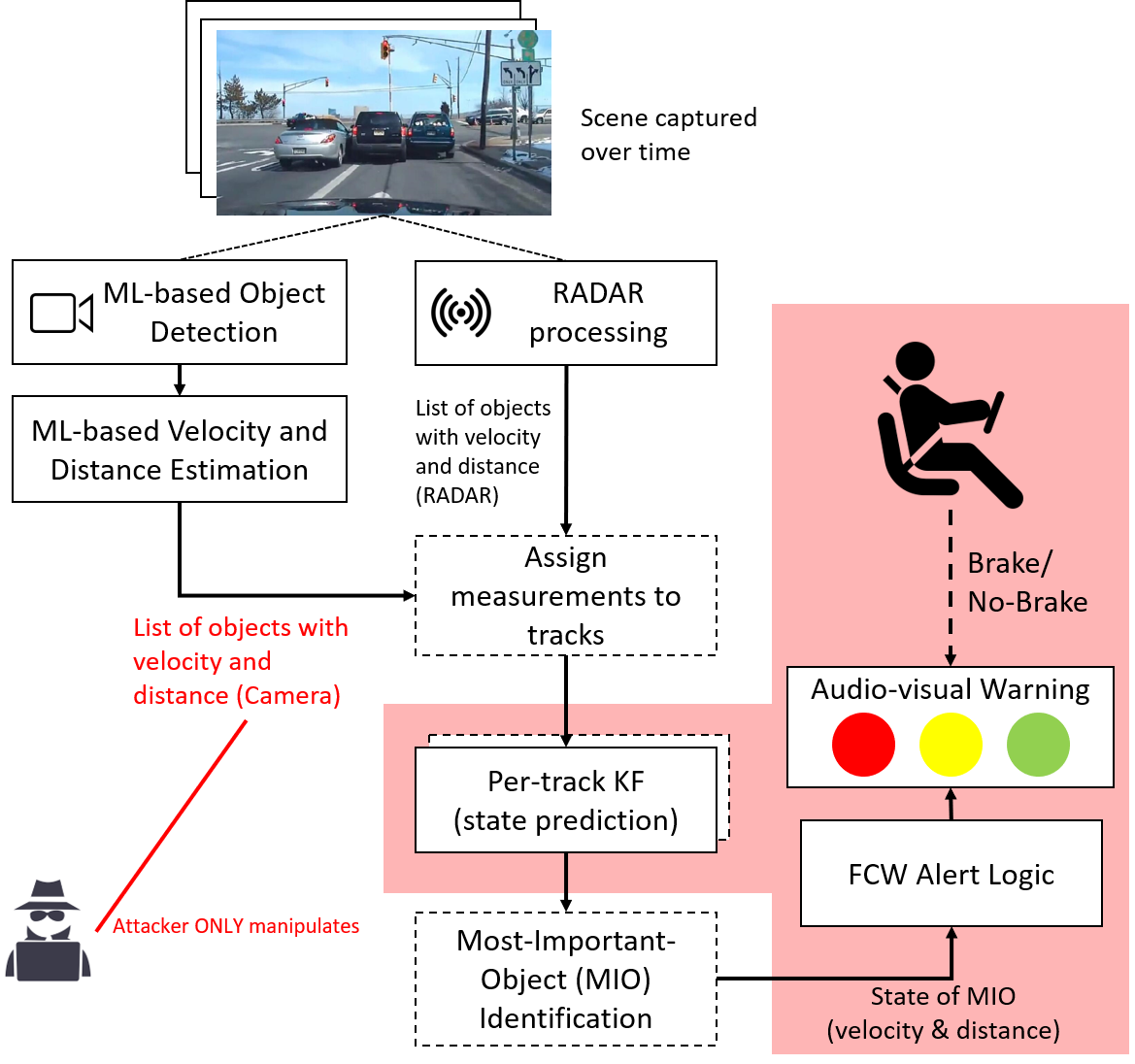}
    \caption{Overview of Forward Collision Warning (FCW) hybrid human-machine system. We take a first step to understanding the robustness of this system to attackers who can compromise sensor measurements. Therefore, we filter the problem to its essence (shaded parts) --- the Kalman filter that tracks the most important object (MIO) and the downstream logic that decides how to warn the driver.}
\label{fig:overview}
\end{figure}

Forward Collision Warning provides audio-visual alerts to warn human drivers of imminent collisions. Fig.~\ref{fig:overview} shows the pipeline of a prototypical FCW hybrid system~\cite{matlab-fcw}: (1) It uses camera and RADAR sensors to perceive the environment; (2) It processes sensor data using a combination of traditional signal processing and machine learning algorithms to derive object velocities and distances; (3) A Kalman filter tracks the Most Important Object (MIO) state and makes predictions about its future states;  (4) FCW logic uses Kalman filter predictions to determine whether a collision is about to occur and creates audio-visual warnings; (5) A human driver reacts to FCW alerts. These alerts can be either: green -- indicating no danger, yellow -- indicating potential danger of forward collision, and red -- indicating imminent danger where braking action must be taken.

We focus on attacking the core steps of FCW (shaded parts of Fig.~\ref{fig:overview}). Thus, we assume there is a single MIO in front of the ego vehicle and a single Kalman filter actively tracking its state. The steps of measurement assignment and MIO identification will not be considered in this paper.

%It keeps track of relevant traffic objects, and determines a unique most important object (MIO) closest to the ego vehicle that can pose danger. FCW uses a Kalman filter to predict the MIO's state in the next timestep and then executes alert logic that uses the predicted state information. The alert logic determines the warning level that FCW surfaces to the driver. Determining the unique MIO among multiple objects on the road involves several algorithmic steps that generally falls under the umbrella of multiple object tracking. Prior work has examined how to trick this step. By contrast, our focus is on attacking the fundamental processing steps of FCW itself and not multiple object tracking. Therefore, we assume that there is a single MIO and that the Kalman filter is already tracking that vehicle.

We have two attack goals that will comprehensively demonstrate the vulnerability of FCW hybrid systems --- the attacker should trick FCW into showing no red alerts when there is an imminent collision with the most important object (MIO), and vice versa --- the attacker should trick FCW into showing red alerts when there is no collision, inducing a human to react with braking that can potentially lead to a rear-end crash with a trailing vehicle.

%Creating the above end-to-end attacks on FCW involves overcoming several challenges: (1) The attack must account for the sequential nature of ADAS where the output (e.g., FCW warning lights) at any point in time also depend on past measurements from both camera and RADAR sensors; (2) At planning time, the attacker does not know future measurements the victim vehicle obtains of the scene; (3) Attacker-induced manipulations to the sensor data cannot be physically implausible (e.g., velocity change of 10,000 m/s) because an outlier detector will flag the change; (4) The attacker must incorporate a model of human reaction to the FCW alerts.

%This procedure is not at the heart of our attack, but could make the problem too much complicated. Therefore, in this paper, we assume there is always a single relevant vehicle, which is the MIO.

\subsection{Kalman Filtering}
At the core of FCW is the Kalman Filter, which estimates the state of the MIO based on sensor measurements. In this paper, the state of the MIO is represented as $x_t = (d_t^1, v_t^1, a_t^1, d_t^2, v_t^2, a_t^2)$, where $d_t^1$, $v_t^1$, $a_t^1$ are the distance, velocity and acceleration of the MIO along the driving direction, and $d_t^2, v_t^2, a_t^2$ for the lateral direction (perpendicular to driving direction). Then KF models the evolution of $x_t$ as
\begin{equation}\label{eq:transition_model}
x_{t+1}=Ax_t + \omega_t, t\ge 1,
\end{equation}
where $A$ is the state-transition matrix and $\omega_t \sim N(0, \Omega)$ is Gaussian noise. 
The underlying state $x_t$ is unknown, but one can obtain measurements $y_t$ of the state as
\begin{equation}\label{eq:measurement_model} 
y_t = C x_t+\psi_t, t\ge 1,
\end{equation}
where $C$ is the measurement matrix and $\psi_t \sim N(0, \Psi)$ is the measurement noise. In our paper, $y_t\in\R^8$ contains vision and radar measurements of the MIO distance and velocity along two directions, i.e., $y_t =(d_t^{1,\nu}, v_t^{1,\nu}, d_t^{2,\nu}, v_t^{2,\nu}, d_t^{1,r}, v_t^{1,r}, d_t^{2,r}, v_t^{2,r})$, where we use superscripts $\nu$, $r$ for vision and radar, and numbers 1, 2 for driving and lateral direction, respectively.
Given the state dynamics~\eqref{eq:transition_model} and measurement model~\eqref{eq:measurement_model}, KF provides a recursive formula to estimate the state based on sequential measurements obtained over time.
Concretely, KF starts from some initial state and covariance prediction $\hat x_1$ and $\hat \Sigma_1$. Then for any $t\ge 2$, KF first applies~\eqref{eq:KF_correction} to correct the predictions based on measurements $y_t$. The corrected state and covariance matrix are denoted by $\bar x_t$ and $\bar \Sigma_t$.
\begin{equation}\label{eq:KF_correction}
\begin{aligned}
\bar x_{t} &= (I-H_{t-1}C)\hat x_{t-1} +H_{t-1} y_t,\\
\bar \Sigma_{t} &= (I - H_{t-1}C)\hat \Sigma_{t-1}.
\end{aligned}
\end{equation}
where $H_{t-1}=\hat \Sigma_{t-1}C^\top (C\hat \Sigma_{t - 1}C^\top+\Psi)^{-1}$.
Next, KF applies~\eqref{eq:KF_prediction} to predict state and covariance for the next step.
\begin{equation}\label{eq:KF_prediction}
\hat x_{t} = A\bar x_{t}, \quad \hat \Sigma_{t}= A\bar \Sigma_{t}A^\top +\Omega.
\end{equation}
The correction and prediction steps are applied recursively as $t$ grows. Note that the derivation of covariance matrix is independent of $y_t$, thus can be computed beforehand.

\subsection{Warning Alert Logic and Human Model}
In this paper, we follow the FCW alert logic used in~\cite{matlab-fcw}.
Let the state prediction be $\hat x_t=(\hat d_t^1, \hat v_t^1, \hat a_t^1, \hat d_t^2, \hat v_t^2, \hat a_t^2)$, then the warning light $\ell_t$ output by FCW at step $t$ is one of the following three cases:
\begin{itemize}
\item Safe (Green): The MIO is moving away, or the distance to MIO remains constant, i.e., $\hat v_{t}^1\ge0$.
\item Caution (Yellow): The MIO is moving closer, but still at a distance further than the minimum safe distance $d^*(\hat v_t^1)$, i.e., $\hat v_{t}^1<0$ and $\hat d_{t}^1>d^*(\hat v_t^1)$. We define the safe distance as $d^*(\hat v_t^1)=-1.2\hat v_{t}^1+(\hat v_{t}^1)^2/0.8g$, where $g$ is 9.8 $m/s^2$.
\item Warn (Red): The MIO is moving closer, and at a distance less than the minimum safe distance, i.e., $\hat v_{t}^1<0$ and $\hat d_{t}^1\le d^*(\hat v_t^1)$.
\end{itemize}
The FCW alert logic can be summarized as:
\begin{equation}\label{eq:FCW}
F(\hat x_{t}) = \left\{
\begin{array}{ll}
\text{green} & \mbox{if $\hat v_{t}^1\ge0$,} \\
\text{yellow} & \mbox{if $\hat v_{t}^1< 0, \hat d_{t}^1>d^*(\hat v_t^1)$,} \\
\text{red} & \mbox{if $\hat v_{t}^1< 0, \hat d_{t}^1\le d^*(\hat v_t^1)$.}
\end{array}
\right.
\end{equation}
Given the FCW warning light, the human driver could be in one of the following two states -- applying the brake pedal, or not applying/releasing the brake. We take into account human reaction time $h^*$; warning lights must sustain at least $h^*$ steps before the human driver switches state. That is, the driver brakes after $h^*$ steps since the first red light, and releases the brake after $h^*$ steps since the first yellow/green light. Note that the yellow and green lights are treated identically in both cases because the MIO is outside the safe distance and no brake is needed. In appendix~\ref{appendix: human}, we provide an algorithmic description of the human model.

\section{Attack Problem Formulation}
We assume white-box setting where the attacker can access the KF parameters (e.g., through reverse engineering). The attacker can directly manipulate measurements (i.e., false data injection), but only pertaining to the vision component, and not the RADAR data. Our attack framework is agnostic of whether the attacker manipulates camera or RADAR, but we choose to only manipulate camera because of the increasing presence of deep learning techniques in ADAS and their general vulnerability to adversarial examples~\cite{szegedy2013intriguing,roadsigns17,athalye2017synthesizing,glasses}. We envision that future work can integrate our results into adversarial examples to create physical attacks.  

We further restrict the attacker to only making physically plausible changes to the vision measurements. This is because an anomaly detection system might filter out physically implausible measurements (e.g., change of $10^4$m/s over one second). Concretely, we require that the distance and velocity measurement after attack must lie in $[\underline d,\bar d]$ and $[\underline v,\bar v]$ respectively. We let $[\underline d, \bar d]=[0,75]$ and $[\underline v,\bar v]=[-30,30]$. Finally, we assume that at any time step, the attacker knows the true measurement only for that time step, but does not know future measurements. To address this difficulty of an unknown future, we propose a model predictive control (MPC)-based attack framework that consists of an outer problem and an inner problem, where the inner problem is an instantiation of the outer problem with respect to attacker-envisioned future in every step of MPC. In the following, we first introduce the outer problem formulation.

\subsection{Outer Attack Problem}
Our attacker has a pre-specified target interval $\T^\dagger$, and aims at changing the warning lights output by FCW in $\T^\dagger$. As a result, the human driver sees different lights and takes unsafe actions. 
Specifically, for any time $t\in \T^\dagger$, the attacker hopes to cause the FCW to output a desired target light $\ell_t^\dagger$, as characterized by~\eqref{attack_fake_sp2:target}, in which $F(\cdot)$ is the FCW alert logic~\eqref{eq:FCW}.
To accomplish this, the attacker manipulates measurements in an attack interval $\T^a$. 
In our paper, we assume $\T^\dagger\subset \T^a$. 
Furthermore, we consider only the scenario where $\T^\dagger$ and $\T^a$ have the same last step, since attacking after the target interval is not needed.
Let $\delta_t$ be the manipulation at step $t$, and $\tilde y_t=y_t+\delta_t$ be measurement after attack.
We refer to the $i$-th component of $\delta_t$ as $\delta_t^i$. 
We next define the attack effort as the cumulative change over measurements $J = \sum_{t\in \T^a}\delta_t^\top R \delta_t$.
where $R\succ 0$ is the effort matrix.
The attacker hopes to minimize the attack effort. 

Meanwhile, the attacker cannot arbitrarily manipulate measurements.
We consider two constraints on the manipulation. 
First, MIO distance and velocity are limited by simple natural physics, as shown in~\eqref{attack_fake_sp2:physical}. 
Moreover, similar to the norm ball used in adversarial examples, we impose another constraint that restricts the attacker's manipulation $\|\delta_t\|_\infty\le \Delta$ (see~\eqref{attack_fake_sp2:delta_max}).
We refer to $\T^s=\T^a\backslash\T^\dagger$, the difference between $\T^a$ and $\T^\dagger$, as the stealthy (or planning) interval. 
During $\T^s$, the attacker can induce manipulations before the target interval with advance planning, and by doing so, hopefully better achieve the desired effect in the target interval. However, for the sake of stealthiness, the planned manipulation should not change the original lights $\ell_t$ during $\T^s$. This is characterized by the stealthiness constraint~\eqref{attack_fake_sp2:stealthy}.

Given all above, the attack can be formulated as:
\begin{eqnarray}
&\min_{\delta_t} &J=\sum_{t\in \T^a}\delta_t^\top R \delta_t,\label{attack_fake_sp2:obj}\\
&\mbox{s.t.} & \tilde y_t=y_t+\delta_t, \forall t\in\T^a,\label{attack_fake_sp2:attack}\\
& & \tilde x_{t}  = A(I-H_{t-1}C)\tilde x_{t-1} +AH_{t-1} \tilde y_t,\label{attack_fake_sp2:KF}\\ 
& & \delta_t^i=0, \forall i\in\I_{\text{radar}}, \forall t\in \T^a,\label{attack_fake_sp2:only_vision}\\
& & \|\delta_t\|\le \Delta, \forall t\in \T^a,\label{attack_fake_sp2:delta_max}\\
& & \tilde d_t^{1,\nu}\in [\underline d,\bar d], \tilde v_t^{1,\nu}\in [\underline v,\bar v], \forall  t\in \T^a,\text{\quad\quad}\label{attack_fake_sp2:physical}\\
& & F(\tilde x_t)=\ell_t^\dagger, \forall t\in \T^\dagger,\label{attack_fake_sp2:target}\\
& & F(\tilde x_t)=\ell_t, \forall t\in \T^s. \label{attack_fake_sp2:stealthy}
\end{eqnarray}
The constraint \eqref{attack_fake_sp2:KF} specifies the evolution of the state prediction under the attacked measurements $\tilde y_t$.~\eqref{attack_fake_sp2:only_vision} enforces no change on radar measurements, where $\I_{\text{radar}}=\{5,6,7,8\}$ contains indexes of all radar components .
The attack optimization is hard to solve due to three reasons: 
\begin{enumerate}[(1).]
\item The problem could be non-convex.
\item The problem could be be infeasible. 
\item The optimization is defined on measurements $y_t$ that are not visible until after $\T^a$, while the attacker must design manipulations $\delta_t$ during $\T^a$ in an online manner. 
\end{enumerate}
We now explain how to address the above three issues.

The only potential sources of non-convexity in our attack are~\eqref{attack_fake_sp2:target} and~\eqref{attack_fake_sp2:stealthy}. We now explain how to derive a surrogate convex problem using $\ell_t^\dagger=\ell_t^o=\text{red}$ as an example. The other scenarios are similar, thus we leave the details to Appendix~\ref{appendix: surrogate_constraint}. The constraint $F(\tilde x_t)=\text{red}$ is equivalent to
\begin{eqnarray}
& \tilde v_t^{1,\nu}&< 0, \label{eq:cons_red_1}\\
& \tilde d_t^{1,\nu}&\le -1.2\tilde v_t^{1,\nu}+\frac{1}{0.8g}(\tilde v_t^{1,\nu})^2.\label{eq:cons_red_2},
\end{eqnarray}
The above constraints result in non-convex optimzation mainly because~\eqref{eq:cons_red_2} is nonlinear. To formulate a convex problem, we now introduce surrogate constraints that are tighter than~\eqref{eq:cons_red_1},~\eqref{eq:cons_red_2} but guarantee convexity.
\begin{proposition}\label{prop:red_surrogate}
Let $U(d)=0.48g-\sqrt{(0.48g)^2+0.8gd}$. Let $\epsilon>0$ be any positive number. Then for any $d_0\ge0$, the surrogate constraints~\eqref{eq:cons_surro_red_1},~\eqref{eq:cons_surro_red_2} are tighter than $F(\tilde x_t)=\text{red}$, and induce convex attack optimization.
\begin{eqnarray}
&\tilde v_t^{1,\nu}&\le -\epsilon,\label{eq:cons_surro_red_1}\\
&\tilde v_t^{1,\nu}&\le U^\prime(d_0)(\tilde d_t^{1,\nu}-d_0)+U(d_0)-\epsilon.\label{eq:cons_surro_red_2}
\end{eqnarray}
\end{proposition}
We provide a proof and guidance on how to select $d_0$ in Appendix~\ref{appendix: surrogate_constraint}. With the surrogate constraints, the attack optimization becomes convex. However, the surrogate optimization might still be infeasible. To address the feasibility issue, we further introduce slack variables into~\eqref{eq:cons_surro_red_1},~\eqref{eq:cons_surro_red_2} to allow violation of stealthiness and target lights:
\begin{eqnarray}
&\tilde v_t^{1,\nu}&\le -\epsilon+\xi_t,\label{eq:cons_surro_slack_red_1}\\
&\tilde v_t^{1,\nu}&\le U^\prime(d_0)(\tilde d_t^{1,\nu}-d_0)+U(d_0)-\epsilon+\zeta_t.\quad\label{eq:cons_surro_slack_red_2}
\end{eqnarray}
We include these slack variables in the objective function:
\begin{equation}\label{eq:definitionJ1J2J3}
J=\underbrace{\sum_{t\in \T^a}\delta_t^\top R \delta_t}_{\text{total manipulation  $J_1$}}+\lambda \underbrace{\sum_{t\in \T^s}(\xi_t^2+\zeta_t^2)}_{\text{stealthiness violation $J_2$}}+\lambda \underbrace{\sum_{t\in \T^\dagger}(\xi_t^2+\zeta_t^2)}_{\text{target violation $J_3$}}.
\end{equation}
Then, the surrogate attack optimization is
\begin{eqnarray}
&\min_{\delta_t} &J=J_1+\lambda J_2+\lambda J_3,\label{attack_surrogate_sp2:obj}\\
&\mbox{s.t.} &\text{\eqref{attack_fake_sp2:attack}-\eqref{attack_fake_sp2:physical},~\eqref{eq:cons_surro_slack_red_1},~\eqref{eq:cons_surro_slack_red_2}}.\label{attack_surrogate_sp2:cons}
\end{eqnarray}
\vspace{-0.5cm}
\begin{proposition}
The attack optimization~\eqref{attack_surrogate_sp2:obj}-\eqref{attack_surrogate_sp2:cons} with surrogate constraints and slack variables is convex and feasible.
\end{proposition}

\subsection{Inner Attack Problem: MPC-based Attack}
In the outer surrogate attack~\eqref{attack_surrogate_sp2:obj}-\eqref{attack_surrogate_sp2:cons}, we need to assume the attacker knows the measurements $y_t$ in the entire attack interval $\T^a$ beforehand. However, the attacker cannot know the future. Instead, he can only observe and manipulate the current measurement in an online manner. To address the unknown future issue, we adopt a control perspective and view the attacker as an adversarial controller of the KF, where the control action is the manipulation $\delta_t$. We then apply MPC, an iterative control method that progressively solves~\eqref{attack_surrogate_sp2:obj}-\eqref{attack_surrogate_sp2:cons}. By using MPC, the attacker is able to adapt the manipulation to the instantiated measurements revealed over time while accounting for unknown future measurements.

Specifically, in each step $t$, the attacker has observed all past measurements $y_{1},...y_{t-1}$ and the current measurement $y_t$. Thus, the attacker can infer the clean state $\hat x_t$ in the case of no attacker intervention. Based on $\hat x_t$, the attacker can recursively predict future measurements by simulating the environmental dynamics without noise, i.e., $\forall  \tau>t$:
\begin{equation}\label{eq:attack_predict}
x_{\tau}^\prime =  A x_{\tau -1}^\prime, \hat y_{\tau} = C x_{\tau}^\prime.
\end{equation}
The recursion starts from $x_t^\prime=\hat x_t$. The attacker then replaces the unknown measurements in the outer attack by its prediction $\hat y_\tau$ $(\tau>t)$ to derive the following inner attack:
\begin{eqnarray}
&\min_{\delta_{\tau:\tau\ge t}} &J=\sum_{\tau\in \T^a}\delta_\tau^\top R \delta_\tau+\lambda \sum_{\tau\in \T^a}(\xi_\tau^2+\zeta_\tau^2),\quad\quad\label{eq:MPC_red_1}\\
&\mbox{s.t.} &\tilde y_\tau=\hat y_\tau+\delta_\tau, \forall \tau\ge t,\\
& & \text{\text{\eqref{attack_fake_sp2:KF}-\eqref{attack_fake_sp2:physical},~\eqref{eq:cons_surro_slack_red_1},~\eqref{eq:cons_surro_slack_red_2}}}\text{ (defined on $\tau\ge t$)} .\label{eq:MPC_cons_red_1}
\end{eqnarray}
The attacker solves the above inner attack in every step $t$. Assume the solution is $\delta_\tau (\tau\ge t)$. Then, the attacker only implements the manipulation on the current measurement, i.e., $\tilde y_t=y_t+\delta_t$, and discards the future manipulations. After that, the attacker enters step $t+1$ and applies MPC again to manipulate the next measurement. This procedure continues until the last step of the attack interval $\T^a$. We briefly illustrate the MPC-based attack in algorithm~\ref{alg:MPC}.
\begin{algorithm}[t]
\SetKwInOut{Input}{Input}\SetKwInOut{Output}{Output}
\Input{target interval $\T^\dagger$, target lights $\ell_t^\dagger, t\in \T^\dagger$, stealthy interval $\T^s$, original lights $\ell_t, t\in \T^s$.}
Initialize $\hat x_1$ and $\hat \Sigma_1$. Let $\tilde x_1=\hat x_1$, $\T^a=\T^s\cup \T^\dagger$\;
\For{$t\leftarrow 2$ \KwTo $T$}{
environment generates measurement $y_t$\;
 \uIf{$t\in\T^a$}
    {
        attacker infers clean state $\hat x_t$ without attack\;
        attacker predicts future $\hat y_t$ with $\eqref{eq:attack_predict}$\;
        attacker solves~\eqref{eq:MPC_red_1}-\eqref{eq:MPC_cons_red_1} to obtain $\delta_\tau$ $(\tau\ge t)$\;
        attacker manipulates $y_t$ to $\tilde y_t=y_t+\delta_t$\;
        $\tilde x_t$ evolves to $\tilde x_{t+1}$ according to $\tilde y_t$
    }
   \lElse
    {
        $\tilde x_t$ evolves to $\tilde x_{t+1}$ according $y_t$
     }
}
 \caption{MPC-based attack.}
 \label{alg:MPC}
\end{algorithm}

% we enforce the edits to the measurement vectors:
% camera distance measurement plus attacker edit should be [0, 80]
% this is because the sensors we have pick up objects at distances upto 80 m but not beyond

% camera velocity measurement plus attacker edit should be [-30, 30] corresponding to 60 mph ego and stationary MIO
% this is physically plausible for example, freeway speeds.

%decel 0.4 g (from our FCW model)
%reaction time 1.2 s (from our FCW model)
%distracted driver time 2 s (from our NHTSA document;  the "eyes off road" stat for 85% of cases in near-crash)
%EGO speed 27 m/s or 60.4 mph (standard freeway speed)
%MIO speed in EGO-MIO collision case: 17 m/s (slowed vehicle on freeway)
%MIO speed in Trailer-EGO collision case: 28 m/s
%Trailer vehicle speed: 27 m/s
%Trailer vehicle following distance: 7 meters (or about 25 feet, average sedan length is 15 feet, so this is more than 1 vehicle length following distance; this is a tailgating vehicle)

\section{Experiments on CARLA Simulation}
In this section, we empirically study the performance of the MPC-based attack. We first describe the simulation setup.

% we enforce the edits to the measurement vectors:
% camera distance measurement plus attacker edit should be [0, 80]
% this is because the sensors we have pick up objects at distances upto 80 m but not beyond

% camera velocity measurement plus attacker edit should be [-30, 30] corresponding to 60 mph ego and stationary MIO
% this is physically plausible for example, freeway speeds.

\subsection{Simulation Setup}
We use CARLA~\cite{carla}, a high-fidelity vehicle simulation environment, to generate measurement data that we input to the Kalman filter-based FCW. CARLA supports configurable sensors and test tracks. We configure the simulated vehicle to contain a single forward-facing RGB camera (800x600 pixels), a forward-facing depth camera of the same resolution, and a single forward-facing RADAR (15$^{\circ}$ vertical detection range, 6000 points/sec, 85 m maximum detection distance). We took this configuration from a publicly-available FCW implementation~\cite{matlab-fcw}. The simulation runs at 20 frames/sec and thus, each sensor receives data at that rate. Furthermore, this configuration is commonly available on production vehicles today~\cite{tesla-adas}, and thus, our simulation setup matches real-world FCW systems from a hardware perspective. 

For each time step of the simulation, CARLA outputs a single RGB image, a depth map image, and variable number of RADAR points. We use YOLOv2~\cite{yolov2} to produce vehicle bounding boxes, the Hungarian pairwise matching algorithm \cite{hungarian} to match boxes between frames, and the first derivative of paired depth map image readings to produce vehicle detections from vision with location and velocity components. Details of processing and formatting of CARLA output can be found in Appendix~\ref{appendix:carla-output-processing}. This process produces measurements that match ground truth velocity and distance closely.

Although there are infinitely many possible physical situations where an FCW alert could occur involving two vehicles, they reside in a small set of equivalence classes. The National Highway Traffic Safety Administration (NHTSA) has outlined a set of testing conditions for assessing the efficacy of FCW alerts~\cite{nhtsa-guidelines}.  It involves a two vehicles on a straight test track at varying speeds. Based on these real-world testing guidelines, we develop the following two scenarios:

\subsubsection{MIO-10: Collision between two moving vehicles}
The ego and MIO travel on a straight road, with a negative relative velocity between the two vehicles. Specifically, the ego travels at 27 m/s (\textasciitilde60 mph) and the MIO at 17 m/s (\textasciitilde38 mph). These correspond to typical freeway speed differences of adjacent vehicles. In the absence of any other action, the ego will eventually collide with the MIO. In our simulations, we let this collision occur and record camera and RADAR measurements throughout. Since the relative velocity of the MIO to the ego is -10m/s, we refer to this dataset as MIO-10.

\subsubsection{MIO+1: No collision}
The ego and MIO travel on a straight road, with a positive relative velocity between the two vehicles. Specifically, the ego travels at 27 m/s (\textasciitilde60 mph) and the MIO at 28 m/s (\textasciitilde63 mph). A trailing vehicle moving at 27 m/s follows the ego 7 m behind.  In the absence of any other action, the ego and trailing vehicle will not collide. We collect measurements until the MIO moves out of sensor range of the ego. We refer to this dataset as MIO+1.

The above scenarios correspond to basic situations where the ego vehicle has an unobstructed view of the MIO and represents a best-case for the FCW system. Attacks on these two settings are the hardest to achieve and comprehensively demonstrate the efficacy of our MPC-based attack.

% we enforce the edits to the measurement vectors:
% camera distance measurement plus attacker edit should be [0, 80]
% this is because the sensors we have pick up objects at distances upto 80 m but not beyond

% camera velocity measurement plus attacker edit should be [-30, 30] corresponding to 60 mph ego and stationary MIO
% this is physically plausible for example, freeway speeds.

\subsection{Attack Setup}
We perform preprocessing of CARLA measurements to remove outliers and interpolate missing data (see Appendix~\ref{appendix: preprocess}). Each step of our KF corresponds to one frame of the CARLA simulated video sequence (i.e., 0.05 seconds). We assume that the KF initializes its distance and velocity prediction to the average of the first vision and RADAR measurements. The acceleration is initialized to 0 in both directions. The covariance matrix is initialized to that used by Matlab FCW \cite{matlab-fcw}. Throughout the experiments, we let the effort matrix $R=I$, the margin parameter $\epsilon=10^{-3}$, and $\lambda=10^{10}$. We assume the human reaction time is $h^*=24$ steps (i.e., 1.2 seconds in our simulation). 

\subsubsection{MIO-10 dataset}
We first simulate FCW to obtain the original warning lights without attack.
The first red light appears at step 98. Before this step, the lights are all yellow.
Without attack, the human driver will notice the red warning at step 98.
After 1.2 seconds of reaction time (24 steps), the driver will start braking at step 122.
The ground-truth distance to the MIO at the first application of brakes is 14.57m.
During braking, the distance between the ego vehicle and the MIO reduces by $10^2/0.8g\approx 12.76$m before stabilizing.
Since this is less than the ground-truth distance of 14.58m before braking, the crash can be avoided. This validates the potential effectiveness of FCW.

Our attacker aims to cause a crash. To accomplish this, the attacker suppresses the first 10 red warnings, so that the first red warning is delayed to step 108. As a result, the driver starts braking at step 132. The ground-truth distance to MIO at this step is 9.58m, which is below the minimum distance needed to avoid collision (12.76m). As such, a collision will occur. Therefore, we let the target interval be $\T^\dagger=[98, 107]$, and the target lights be $\ell_t^\dagger=\text{green}, \forall t\in \T^\dagger$.

\subsubsection{MIO+1 dataset}
In this scenario, the original warning lights without attack are all green. There is a trailing vehicle 7 m behind the ego vehicle, driving at the same velocity as the ego vehicle. Our attacker aims at causing the FCW to output red lights, so that the ego vehicle suddenly brakes unnecessarily and causes a rear collision with the trailing vehicle. To this end, the attacker changes the green lights in the interval [100, 139] to red, in which case the ego vehicle driver starts braking at step 124, after 1.2 seconds of reaction time. If the warning returns to green at step 140, the driver will react after 1.2 seconds and stop braking at step 164. Therefore, the driver continuously brakes for at least $(164-124)\times0.05=2$ seconds. Assuming the driver of the trailing vehicle is distracted, then during those 2 seconds, the distance between the trailing and the ego vehicle reduces by $0.2g \times2^2=7.84m>7m$, thus causing a rear-collision. Therefore, we let the target interval be $\T^\dagger=[100,139]$ and the target lights be $\ell_t^\dagger=\text{red}, \forall t\in \T^\dagger$.

\begin{figure*}[t]
\centering
	 \begin{subfigure}{.24\textwidth}
		\centering
		\includegraphics[width=1\textwidth, height=0.7\textwidth]{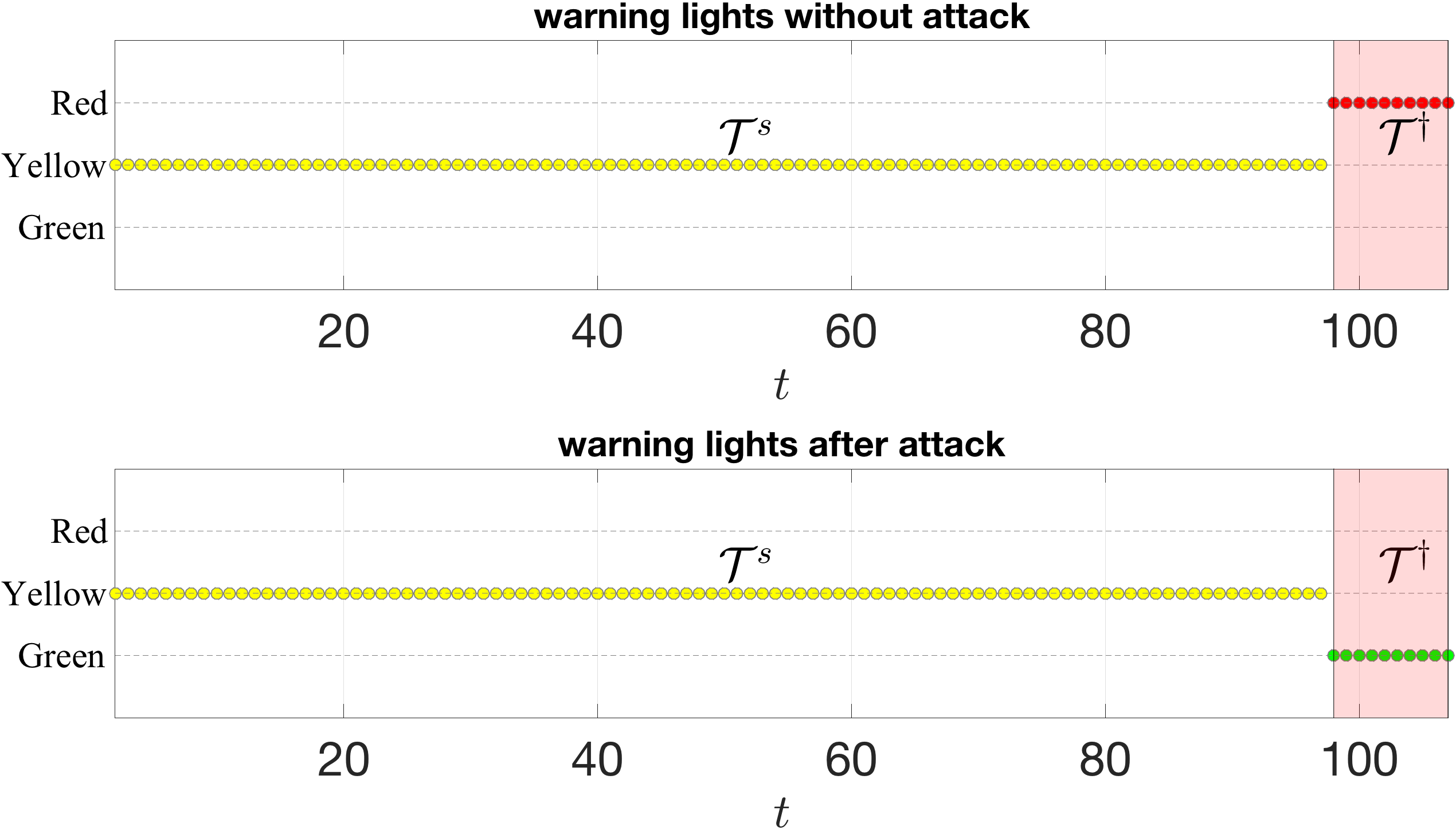}
		\caption{The warning lights.}
		\label{fig:MIO-10_warning}
	\end{subfigure}%
	\hfill
	\begin{subfigure}{.24\textwidth}
		\centering
		\includegraphics[width=1\textwidth, height=0.7\textwidth]{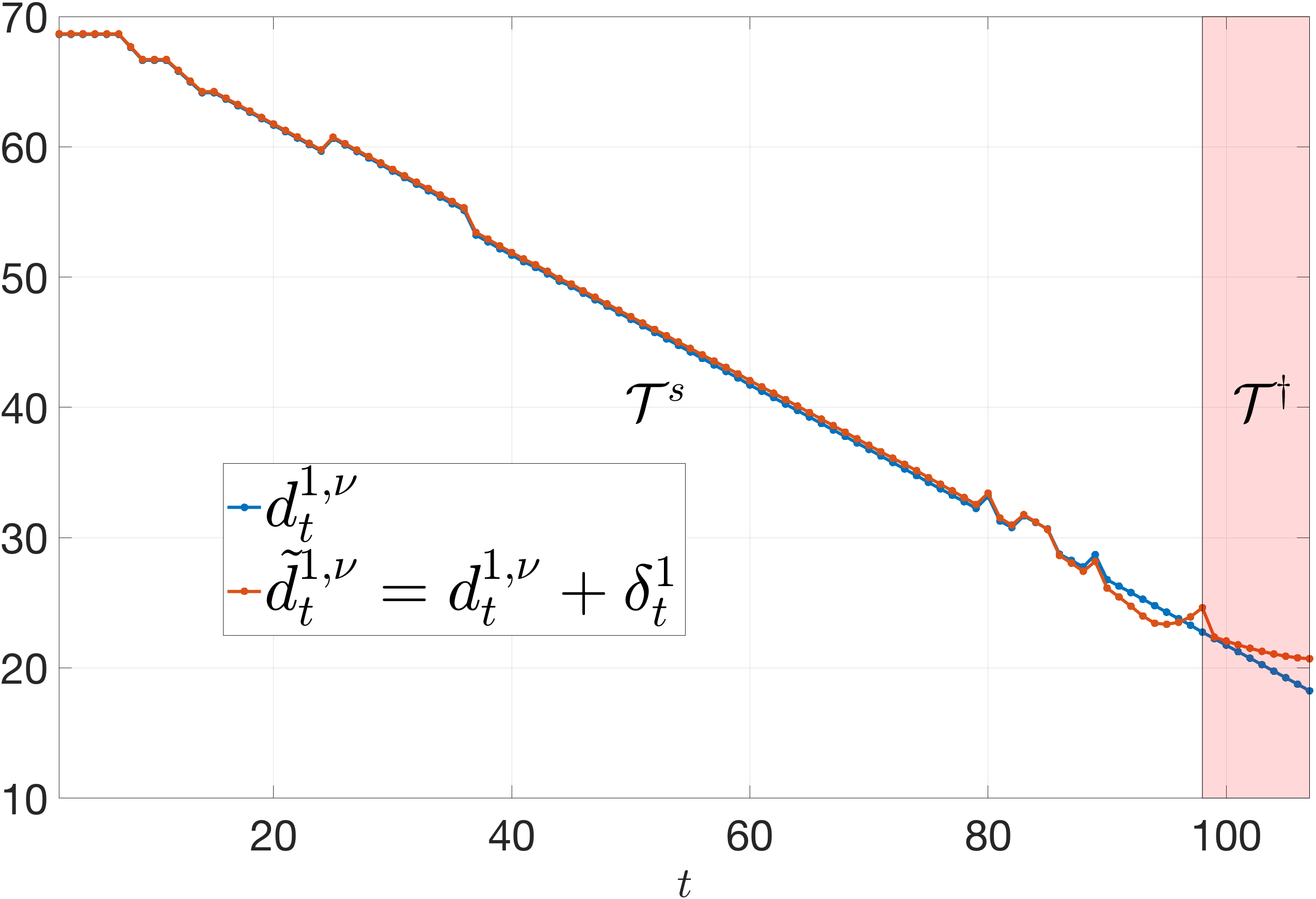}
		\caption{The manipulation on distance.}
		\label{fig:MIO-10_attack_distance}
	\end{subfigure}%
	\hfill
	\begin{subfigure}{.24\textwidth}
		\centering
		\includegraphics[width=1\textwidth, height=0.7\textwidth]{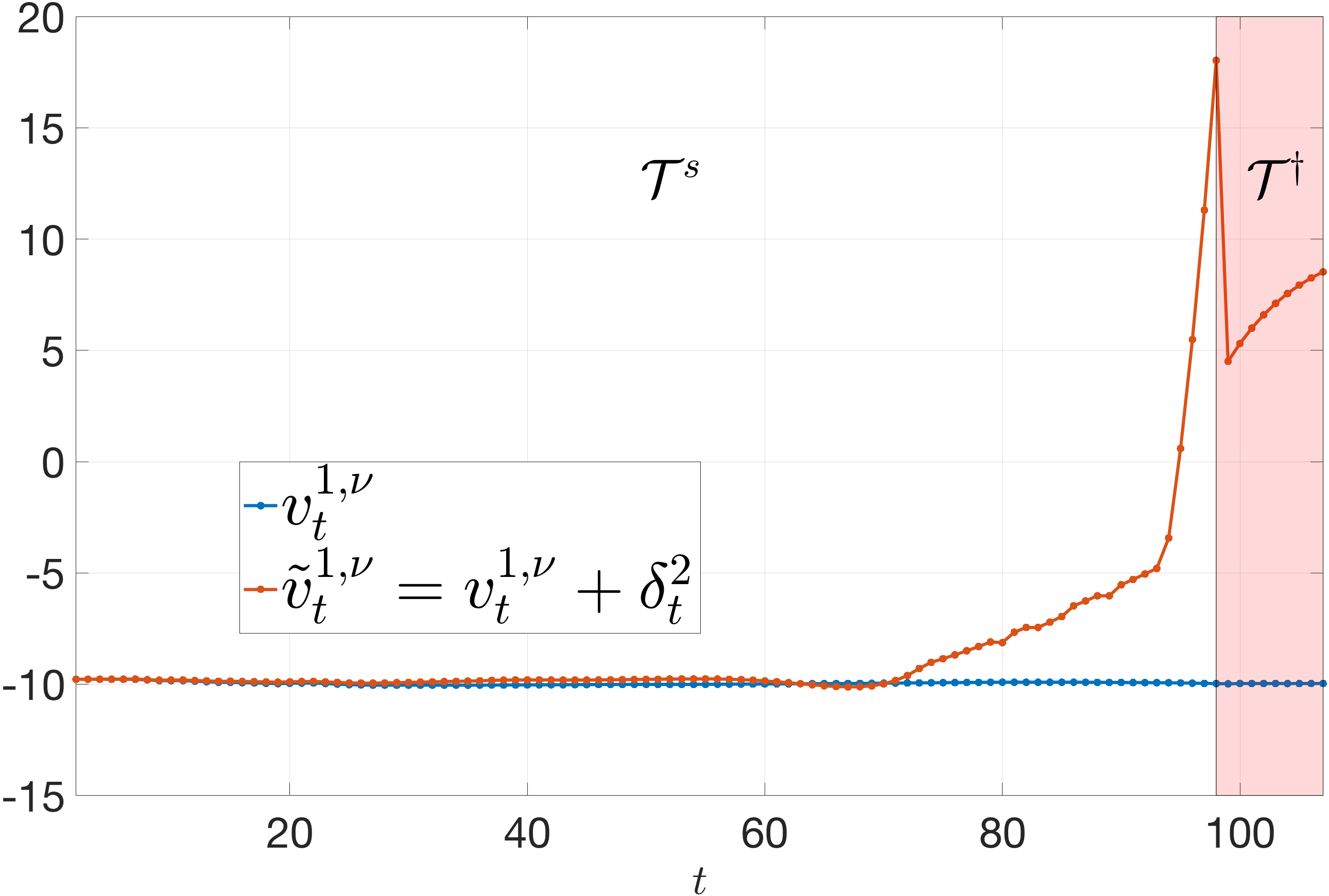}
		\caption{The manipulation on velocity.}
		\label{fig:MIO-10_attack_velocity}
	\end{subfigure}%
	\hfill
	\begin{subfigure}{.24\textwidth}
		\centering
		\includegraphics[width=1\textwidth, height=0.7\textwidth]{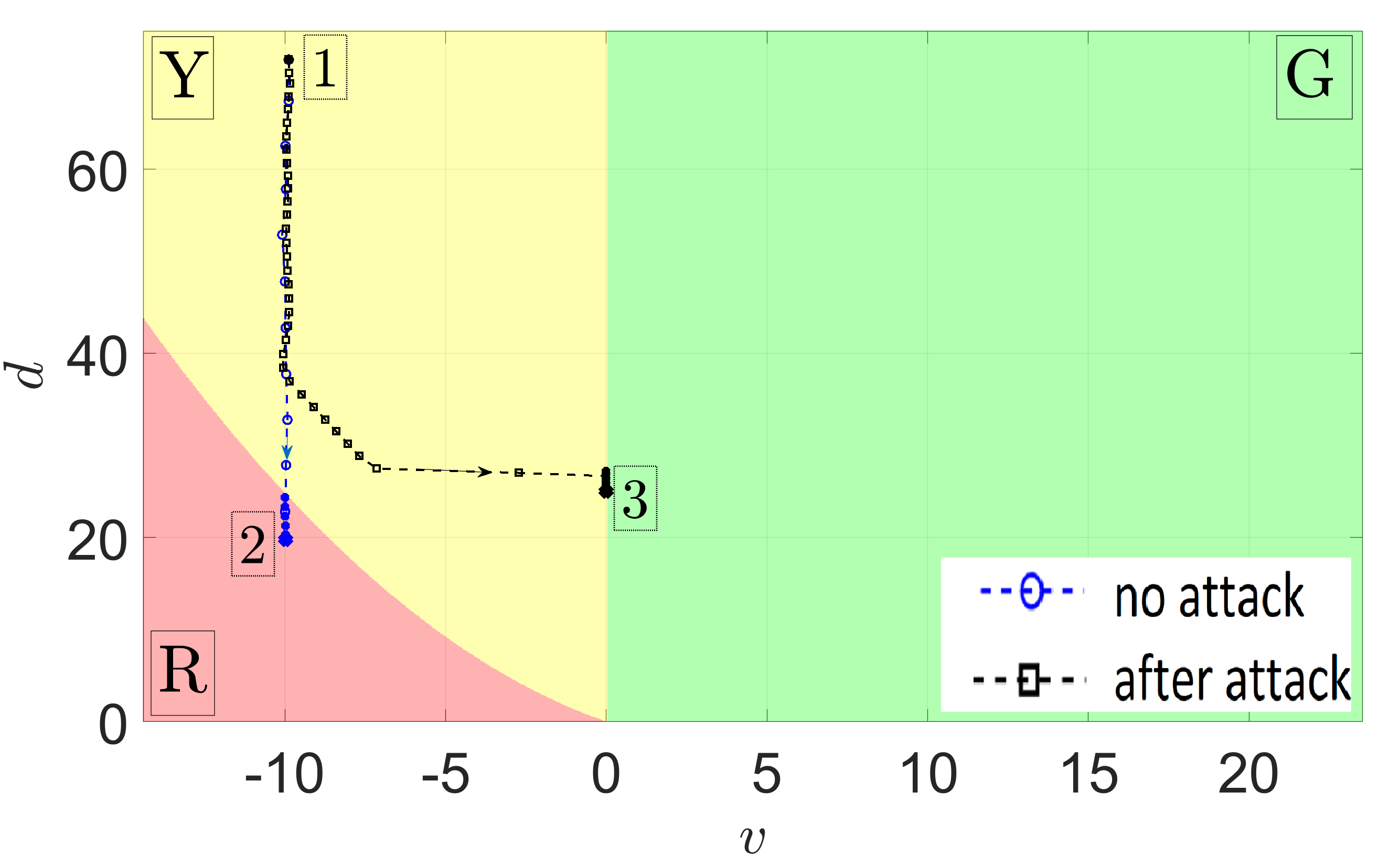}
		\caption{The state trajectory.}
		\label{fig:MIO-10_state_traj}
	\end{subfigure}%
	\caption{Attacks on the MIO-10 dataset.}
	\label{fig:attack_MIO-10}
\end{figure*}

\begin{figure*}[t]
	 \begin{subfigure}{.24\textwidth}
		\centering
		\includegraphics[width=1\textwidth, height=0.7\textwidth]{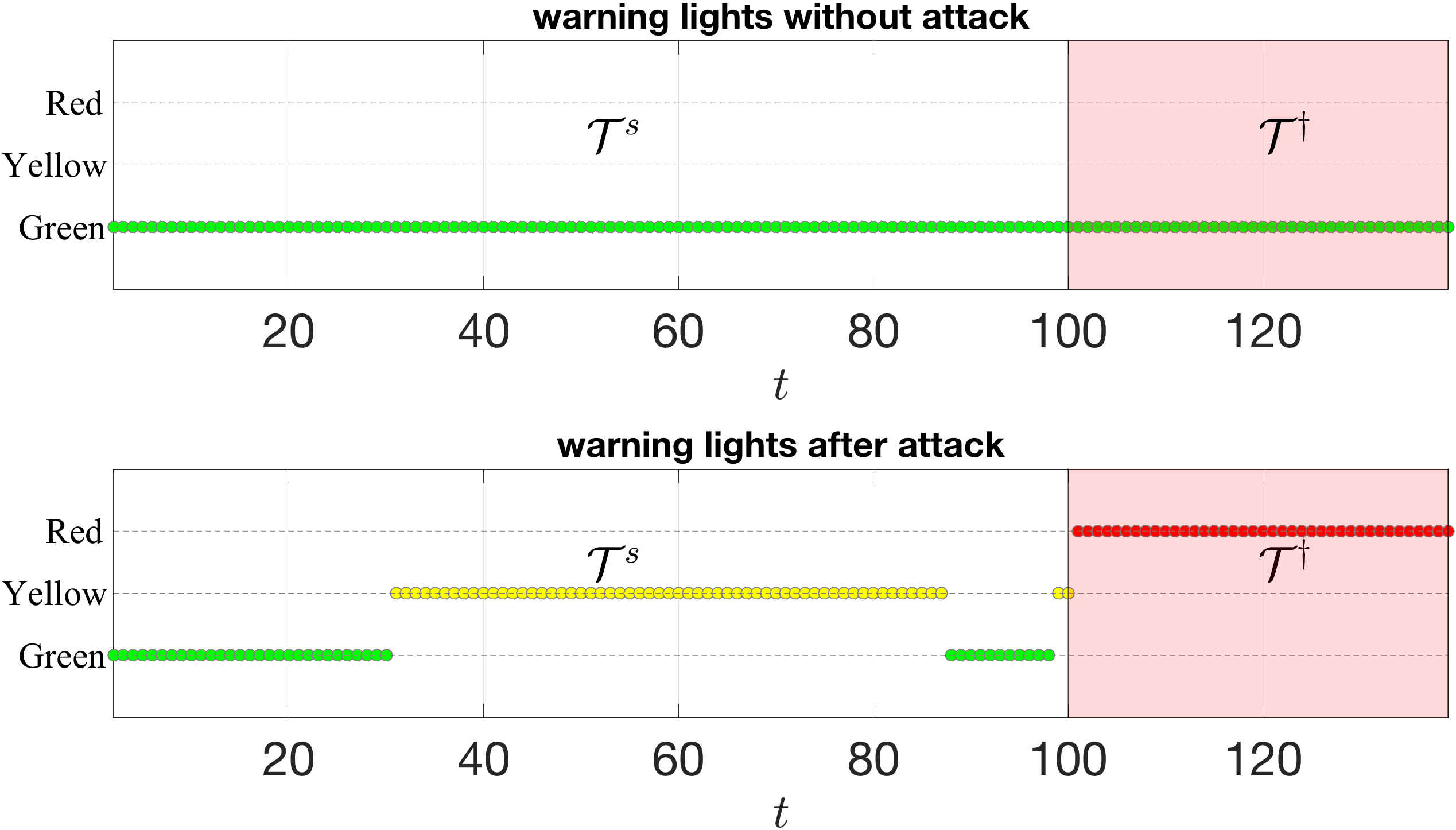}
		\caption{The warning lights.}
		\label{fig:MIO+1_warning}
	\end{subfigure}%
	\hfill
	\begin{subfigure}{.24\textwidth}
		\centering
		\includegraphics[width=1\textwidth, height=0.7\textwidth]{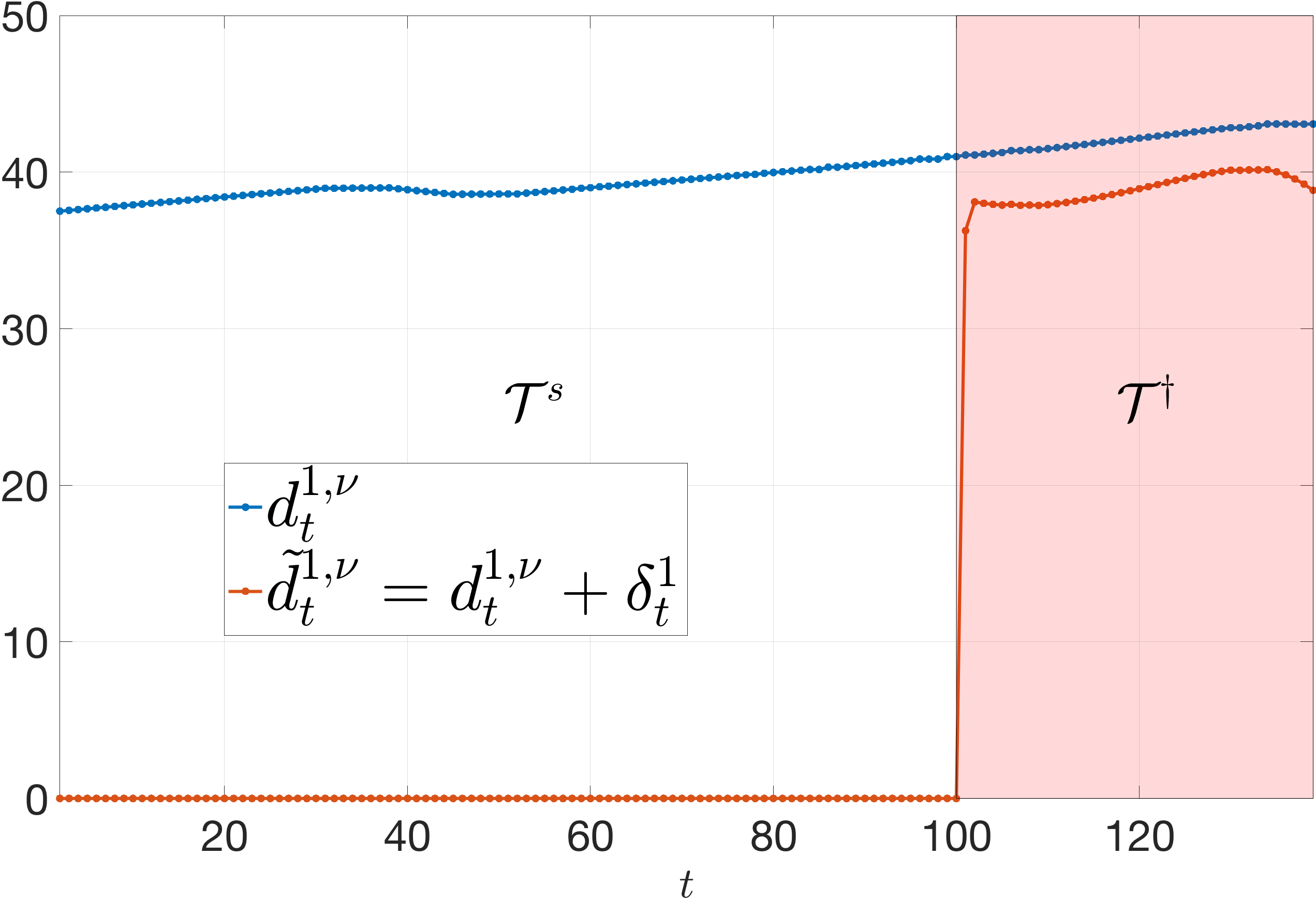}
		\caption{The manipulation on distance.}
		\label{fig:MIO+1_attack_distance}
	\end{subfigure}%
	\hfill
	 \begin{subfigure}{.24\textwidth}
		\centering
		\includegraphics[width=1\textwidth, height=0.7\textwidth]{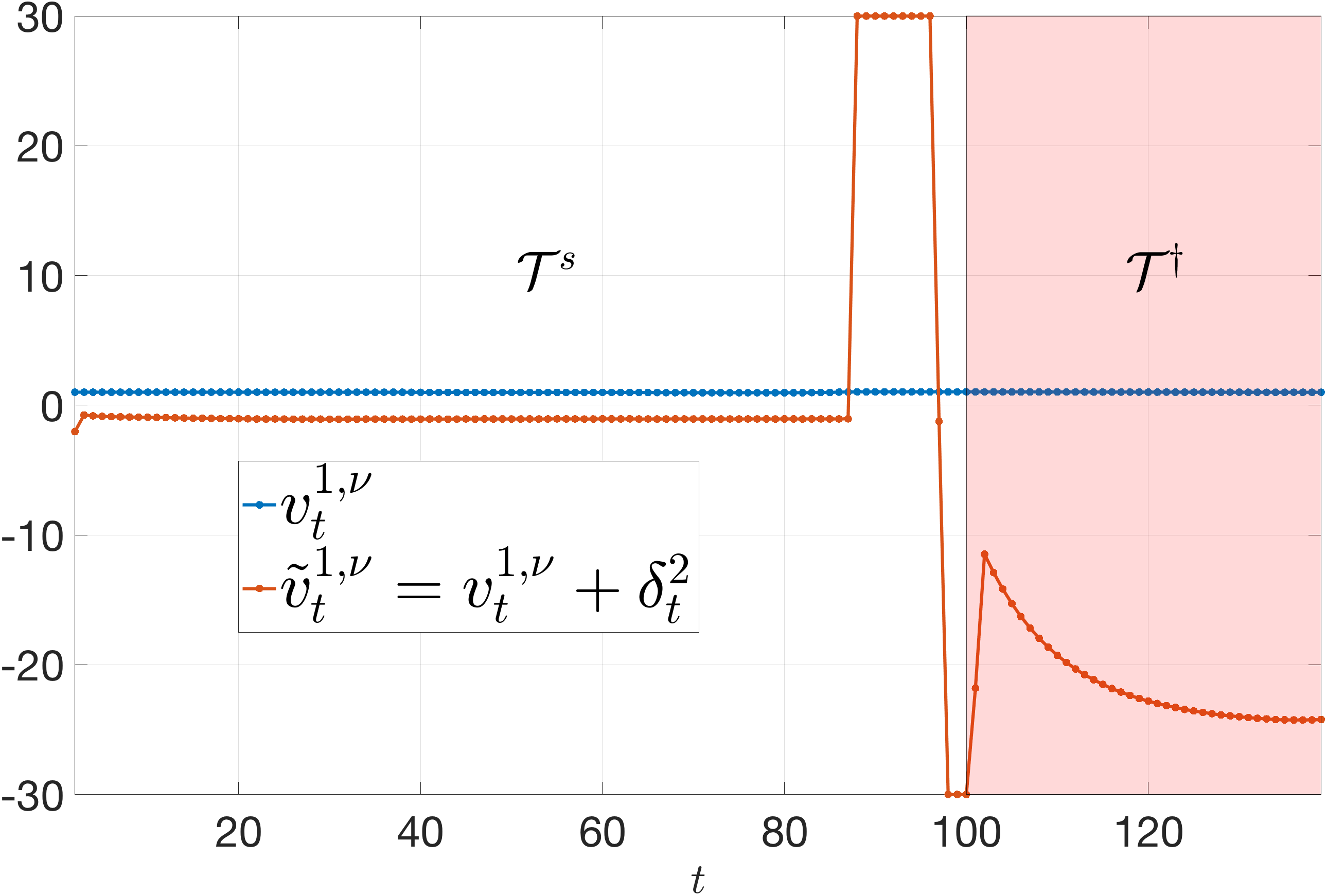}
		\caption{The manipulation on velocity.}
		\label{fig:MIO+1_attack_velocity}
	\end{subfigure}%
	\hfill
	\begin{subfigure}{.24\textwidth}
		\centering
		\includegraphics[width=1\textwidth, height=0.7\textwidth]{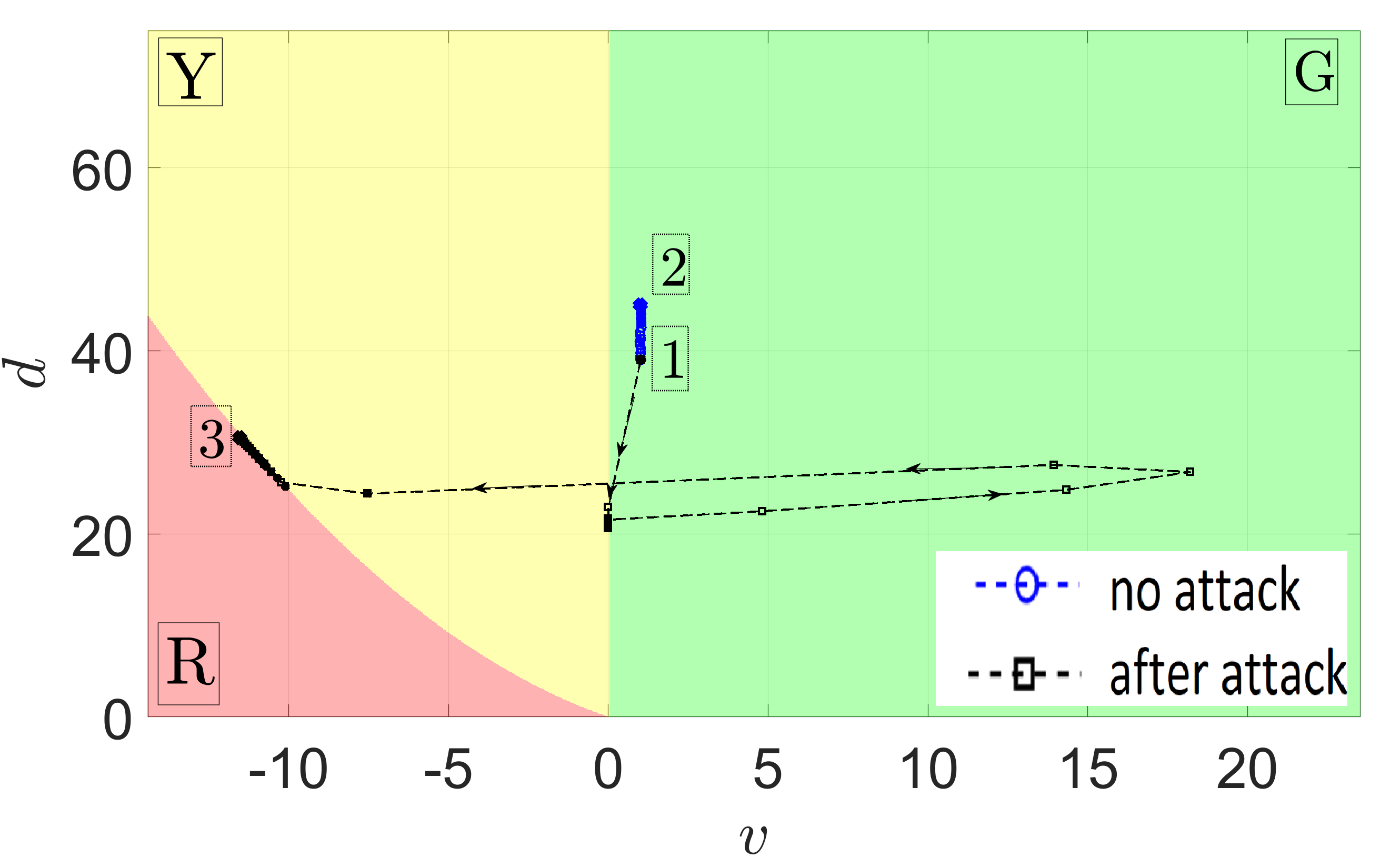}
		\caption{The state trajectory.}
		\label{fig:MIO+1_state_traj}
	\end{subfigure}%
	\caption{Attacks on the MIO+1 dataset.}
	\label{fig:attack_MIO+1}
\end{figure*}

\subsection{The MPC-based Attack Is Successful}
Our first result shows that the MPC-based attack can successfully cause the FCW to output the desired warning lights in the target interval $\T^\dagger$. In this experiment, we let $\Delta=\infty$ and the stealthy interval $\T^s$ start at step 2. In Fig.~\ref{fig:MIO-10_warning} and~\ref{fig:MIO+1_warning}, we show the warning lights in $\T^\dagger$ (shaded in red). For MIO-10, the attacker achieves the desired red lights in the entire $\T^\dagger$, while maintaining the original yellow lights in $\T^s$. For MIO+1, the attacker failed to achieve the red warning at step 100, but is successful in all later steps. We verified that the attack still leads to a collision. In fact, the attacker can tolerate at most two steps of failure in the beginning of $\T^\dagger$ while still ensuring that the collision occurs. There is an unintended side effect in $\T^s$ where green lights are changed to yellow. However, this side effect is minor since the driver will not brake when yellow lights are produced. In many production vehicles, green and yellow lights are not shown to the driver --- only the red warnings are shown.

In Fig.~\ref{fig:MIO-10_attack_distance},~\ref{fig:MIO-10_attack_velocity}, we note that for MIO-10, the manipulation is mostly on velocity, and there are early planned manipulations starting from step 70. A large increase in velocity happens at step 100 (the first step of $\T^\dagger$), which causes the KF's velocity estimation to be positive, resulting in a green light. After that, velocity measurements are further increased to maintain a positive velocity estimation. In Fig.~\ref{fig:MIO+1_attack_distance},~\ref{fig:MIO+1_attack_velocity}, we show manipulations on MIO+1. The overall trend is that the attacker reduces the perceived MIO distance and velocity. As a result, KF estimates the MIO to be close than the safe distance in $\T^\dagger$, thus red lights are produced. During interval [88,96], There is an exceptional increase of velocity. We provide a detailed explanation for that increase in Appendix~\ref{appendix:increase}.

In Fig.~\ref{fig:MIO-10_state_traj},~\ref{fig:MIO+1_state_traj}, we show the trajectory of KF state prediction projected onto the distance-velocity space during interval $\T^a$. We partition the 2D space into three regions, green (G), yellow (Y) and red (R). Each region contains the states that trigger the corresponding warning light. The trajectory without attack (blue) starts from location 1 and ends at 2. After attack, the trajectory (dark) is steered into the region of the desired warning light, ending at location 3. Note that during $\T^\dagger$, the state after attack lies on the boundary of the desired region. This is because our attack minimizes manipulation effort. Forcing a state deeper into the desired region would require more effort, increasing the attacker's cost.

\subsection{Attack Is Easier with More Planning Space}
Our second result shows that the attack is easier when the attacker has more time to plan, or equivalently, a longer stealthy interval $\T^s$. The stealthy interval is initially of full length, which starts from step 2 until the last step prior to $\T^\dagger$. Then, we gradually reduce the length by 1/4 of the full length until the interval is empty. This corresponds to 5, 3.75, 2.5, 1.25 and 0 seconds of planning space before the target interval $\T^\dagger$. We denote the number of light violations in $\T^\dagger$ as $V^\dagger=\sum_{t\in \T^\dagger}\tilde \ell_t\neq \ell_t^\dagger$, and similarly $V^s$ for $\T^s$. We let $\Delta=\infty$. In Table~\ref{table:MIO-10} and~\ref{table:MIO+1}, we show $V^\dagger$, $V^s$ together with $J_1, J_2, J_3$ and $J$ as defined in~\eqref{eq:definitionJ1J2J3} for MIO-10 and MIO+1 respectively. Note that on both datasets, the violation $V^\dagger$ and the total objective $J$ decrease as the length of $\T^s$ grows, showing that the attacker can better accomplish the attack goal given a longer interval of planning. 

On MIO-10, when $\T^s$ is empty, the attack fails to achieve the desired warning in all target steps. However, given 1.25s of planning before $\T^\dagger$, the attacker forces the desired lights throughout $\T^\dagger$. Similarly, on MIO+1, when $\T^s$ is empty, the attack fails in the first three steps of $\T^\dagger$, and the collision will not happen. Given 1.25s of planning before $\T^\dagger$, the attack only fails in the first step of $\T^\dagger$, and the collision happens. This demonstrates that planning in $\T^s$ benefits the attack.
\begin{table}
\centering
\caption{\small{$V^\dagger$, $V^s$, $J_1$, $J_2$, $J_3$ and $J$ for the MIO-10 dataset.}}\label{table:MIO-10}
\small
\resizebox{0.48\textwidth}{!}{
	\begin{tabular}{|c|c|c|c|c|c|c|c|c|c|c|c|c|} 
		\hline
		&  \multicolumn{6}{|c|}{MPC-based attack} &  \multicolumn{6}{|c|}{Greedy attack} \\
		\hline
		$\T^s$  &$V^\dagger$ & $V ^s$ & $J_1$ &  $J_2$  &  $J_3$  &  $J$ &$V^\dagger$ &$V^s$  &  $J_1$ &  $J_2$  &  $J_3$  &  $J$ \\ 
		\hline
		$0$ & 1   & 0 &7.1e3 &0 &7.4 &7.4e10 & 1 & 0  &4.6e3 &98.4 &7.4 &1.1e12 \\ 
		$1.25$ &0  & 0&4.4e3&0 &0&4.3e3  & 0  & 23&1.3e5 &3.3e3 &0 & 3.3e13 \\
		$2.5$ &0 & 0 &4.4e3&0 &0 & 4.4e3  & 0  & 47 &2.0e5 &5.4e3 &0 &5.4e13 \\
		$3.75$ &0 & 0 &4.4e3&0 &0 & 4.4e3 & 0  & 71 &2.5e5 &7.6e3 &0  &7.5e13\\
		$5$ & 0 & 0 &4.4e3&0 &0 & 4.4e3 & 0  & 96 & 2.9e5  &9.2e3 & 0& 9.2e13 \\
		\hline
	\end{tabular}}
\end{table}
\begin{table}
\centering
\caption{\small{$V^\dagger$, $V^s$, $J_1$, $J_2$, $J_3$ and $J$ for the MIO+1 dataset.}}\label{table:MIO+1}
\small
\resizebox{0.48\textwidth}{!}{
	\begin{tabular}{|c|c|c|c|c|c|c|c|c|c|c|c|c|} 
		\hline
		&  \multicolumn{6}{|c|}{MPC-based attack} &  \multicolumn{6}{|c|}{Greedy attack} \\
		\hline
		$\T^s$  &$V^\dagger$ & $V ^s$ &  $J_1$ &  $J_2$  &  $J_3$  &  $J$ &$V^\dagger$ & $V ^s$ &  $J_1$ &  $J_2$  &  $J_3$  &  $J$ \\ 
		\hline
		$0$ & 3 & 0 &3.3e4 &0 &1.2e2 &1.2e12 & 3 & 0  &1.1e5 &0 &1.2e2 &1.2e12 \\ 
		$1.25$ &1 & 14 &7.6e4&6.8 &11.0&1.8e11  & 0 & 25 &1.7e5 &6.1e3 &0 & 6.1e13 \\
		$2.5$ &1& 39 &1.1e5&4.2 &6.9 & 1.1e11 & 0 & 49 &2.3e5 &1.1e4 &0 &1.1e14 \\
		$3.75$ &1 & 58 &1.5e5&3.5 &5.9 & 9.4e10 & 0 & 74 &3.0e5 &1.6e4 &0  &1.6e14\\
		$5$ & 1 & 58 &1.8e5&3.3 &5.6 & 9.0e10 & 0 & 98 & 3.5e5  &2.0e4 & 0& 2.0e14 \\
		\hline
	\end{tabular}}
\end{table}

\subsection{Attack Is Easier as $\Delta$ Increases}
\begin{figure}[H]
	\begin{subfigure}{.235\textwidth}
		\centering
		\includegraphics[width=1\textwidth, height=0.7\textwidth]{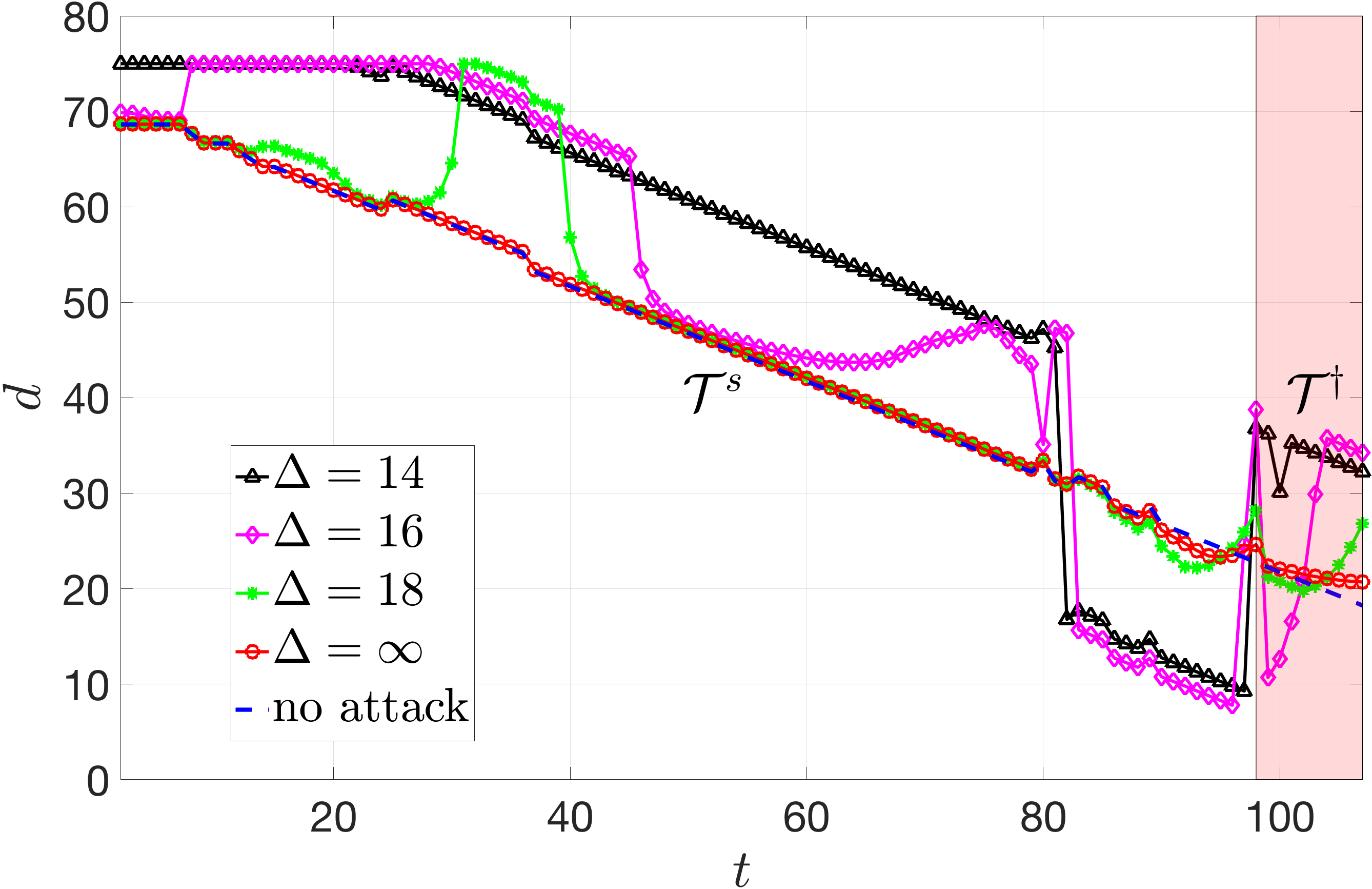}
		\caption{Manipulation on distance.}
		\label{fig:MIO-10_distance_Delta}
	\end{subfigure}%
	\hfill
	 \begin{subfigure}{.235\textwidth}
		\centering
		\includegraphics[width=1\textwidth, height=0.7\textwidth]{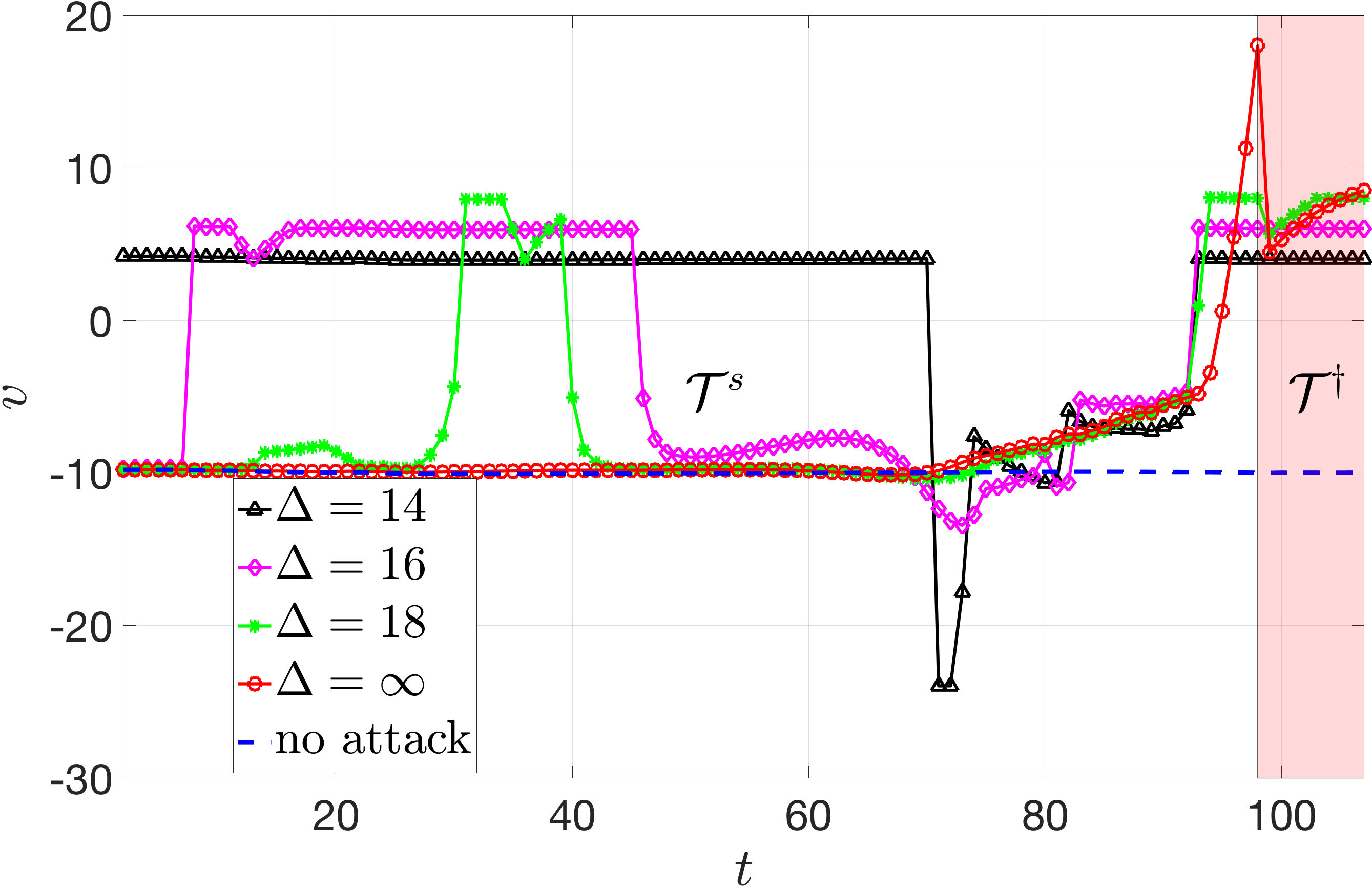}
		\caption{Manipulation on velocity.}
		\label{fig:MIO-10_velocity_Delta}
	\end{subfigure}%
	\caption{Manipulation on measurements with different upper bound $\Delta$. As $\Delta$ grows, the attack becomes easier.}
	\label{fig:manipulation_MIO-10_delta}
\end{figure}

In this section, we show that the attack becomes easier as the upper bound on the manipulation $\Delta$ grows. In this experiment, we focus on the MIO-10 dataset and let $\T^\dagger$ start from step 2.  In Fig~\ref{fig:manipulation_MIO-10_delta}, we show the manipulation on measurements for $\Delta=14,16,18$ and $\infty$. The number of green lights achieved by the attacker in the target interval is 0, 4, 10 and 10 respectively. This shows the attack is easier for larger $\Delta$. Note that for smaller $\Delta$, the attacker's manipulation becomes flatter due to the constraint $\|\delta_t\|\le \Delta$. But, more interestingly, the attacker needs to start the attack earlier to compensate for the decreasing bound.  We also note that the minimum $\Delta$ to achieve the desired green lights over the entire target interval (to integer precision) is 18. 

\subsection{Comparison Against Greedy Attacker}
In this section, we introduce a greedy baseline attacker. For MIO-10, since the attack goal is to achieve green lights in $\T^\dagger$, the greedy attacker always increases the distance and velocity to the maximum possible value, i.e., 
$$\tilde d_t^{1,\nu}=\min\{d_t^{1, \nu}+\Delta, \bar d\}, \tilde v_t^{1,\nu}=\min\{v_t^{1, \nu}+\Delta, \bar v\}, \forall t\in \T^a.$$
Similarly, for MIO+1, the attacker always decreases the distance and velocity to the minimum possible value.

In table~\ref{table:MIO-10} and~\ref{table:MIO+1}, we compare the performance of greedy and our MPC-based attack. On both datasets, each attack strategy achieves a small number of violations $V^\dagger$ in $\T^\dagger$. However, the greedy attack suffers significantly more violations $V^s$ in $\T^s$ than does MPC. Furthermore, these violations are more severe, reflected by the much larger $J_2$ of the greedy attack. As an example, on MIO+1, the greedy attack changes the original green lights in $\T^s$ to red, while our attack only changes green to yellow. The greedy attack also results in larger total effort $J_1$ and objective value $J$. Therefore, we conclude that our attack outperforms the baseline greedy attack overall. In appendix~\ref{appendix: greedy}, we provide more detailed results of the greedy attack.

% we enforce the edits to the measurement vectors:
% camera distance measurement plus attacker edit should be [0, 80]
% this is because the sensors we have pick up objects at distances upto 80 m but not beyond

% camera velocity measurement plus attacker edit should be [-30, 30] corresponding to 60 mph ego and stationary MIO
% this is physically plausible for example, freeway speeds.

%\item Discuss related works such as that ICLR paper.

%\item Discuss the false data injection attacks on Kalman Filter in traditional control community, and articulate the difference.

%\item Data poisoning attacks and machine teaching.

%\item 

\section{Related Work}

\noindent\textbf{Attacks on Object Tracking.} Recent work has examined the vulnerability of multi-object tracking (MOT)~\cite{iclr-mot}. Although this work does consider the downstream logic that uses the outputs of ML-based computer vision, our work goes beyond in several ways. First, we consider a hybrid system that involves both human and machine. Second, we consider the more realistic case of sensor fusion involving RADAR and camera measurements that is deployed in production vehicles today. Prior work assumed a system that only uses a single camera sensor. Third, we examine a complete FCW pipeline that uses object tracking data to predict collisions and issue warnings. Prior work only considered MOT without any further logic that is necessarily present in realistic systems. Finally, our attack algorithm accounts for the sequential nature of decision making in ADAS. 

\noindent\textbf{Vision Adversarial Examples.} ML models are vulnerable to adversarial examples~\cite{szegedy2013intriguing}, with a bulk of research in the computer vision space~\cite{goodfellow2014explaining,papernot2016limitations,carlini2017towards,shafahi2018poison,chen2017targeted}. Recent work has demonstrated physical attacks in the real world~\cite{patch,athalye2017synthesizing,yolo,glasses}. For example, attackers can throw inconspicuous stickers on stop signs and cause the model to output a speed limit sign~\cite{roadsigns17}. However, all of this work studies the ML model in isolation without considering the cyber-physical system that uses model decisions. By contrast, we contribute the first study that examines the security of FCW --- a hybrid human-machine system, and we introduce a novel control-based attack that accounts for these aspects while remaining stealthy to the human driver.

\noindent\textbf{Control-based Attacks on KF.} Prior work in control theory has studied false data injection attacks on Kalman filters~\cite{bai2017kalman, kung2016performance, zhang2016stealthy, chen2016cyber, yang2016false, chen2017optimal}. Our work assumes a similar attack modality -- the attacker can manipulate measurements. However, prior work does not consider the downstream logic and human behavior that depends on the KF output. By contrast, we provide an attack framework demonstrating end-to-end effects that cause crashes in distracted driving scenarios.

\noindent\textbf{Attacks on Sequential Systems.} There are recent works that study attacks of other sequential learning systems from a control perspective~\cite{chen2020optimal,zhang2020online,zhang2020adaptive,jun2018adversarial}. Most of them focus on analyzing theoretical attack properties, while we contribute an application of control-based attacks in a practical domain.

%However, these attacks are designed against KF alone as a single theoretical model without considering the logics used in downstream applications. Our paper distinguishes in that we treat KF as part of the more complex FCW-human system, and we aim at designing end-to-end attacks that can change the output of the system (human behavior) as a whole, thus is more practically relevant.

% we enforce the edits to the measurement vectors:
% camera distance measurement plus attacker edit should be [0, 80]
% this is because the sensors we have pick up objects at distances upto 80 m but not beyond

% camera velocity measurement plus attacker edit should be [-30, 30] corresponding to 60 mph ego and stationary MIO
% this is physically plausible for example, freeway speeds.

\section{Conclusion}
We formulate the adversarial attack of Kalman Filter as an optimal control problem. We demonstrate that our planning-based attack can manipulate the FCW to output incorrect warnings, which mislead human drivers to behave unsafely and cause crash. Our study incorporates human behaviors and applies to general machine-human hybrid systems.

\section{Acknowledgments}
This work was supported in part by the University of Wisconsin-Madison Office of the Vice Chancellor for Research and Graduate Education with funding from the Wisconsin Alumni Research Foundation. XZ acknowledges NSF grants 1545481, 1704117, 1836978, 2041428, 2023239 and MADLab AF CoE FA9550-18-1-0166.

\section{Ethics Statement}
Our paper studies attacks on advanced driver assistance systems (ADAS) with the goal of initiating research into defenses. We do not intend for the attacks to be deployed in the real world. However, studying attacks is critical to understanding what types of defenses must be built and where defense efforts should be focused. We take a first step towards robust ADAS by studying attacks on Kalman filters that are popularly used in these systems.

\bibliography{ref}

%Our ultimate goal is not to deploy the attacks in real wrold, but instead inspire researchers in the adversarial machine learning community to think about how to design defense mechanisms against our attack. To this end, we take a first step towards understanding the vulnerability of the ADAS, especially with respect to its core component ? Kalman Filter. We hope our study can trigger more research along the defense direction.
\clearpage
\appendix
\section{Simulated Raw Data Processing} \label{appendix:carla-output-processing}
CARLA outputs a single RGB image, a depth map image, and variable number of RADAR points for each 0.05 second time step of the simulation. We analyze this data at each time step to produce object detections in the same format of MATLAB FCW \cite{matlab-fcw}:

\[
\begin{bmatrix}
    d^1 & v^1 & d^2 & v^2
\end{bmatrix}
\]

where $d^1$ and $d^2$ are the distance, in meters, from the vehicle sensor in directions parallel and perpendicular to the vehicle's motion, respectively. $v^1$ and $v^2$ are the detected object velocities, in m/s, relative to the ego along these parallel and perpendicular axes.

To produce these detections from vision data, we first find bounding boxes around probable vehicles in each RGB image frame using an implementation of a YOLOv2 network in MATLAB which has been pre-trained on vehicle images \cite{matlab-yolo}. Each bounding box is used to create a distinct object detection. The $d^1$ value, or depth, of each object is taken to be the depth recorded by the depth map at the center pixel of each bounding box.

The $d^2$ value of each detection is then computed as

\begin{equation}
d^2 = u * \frac{d^1}{l_{foc}}
\end{equation}

where $u$ is the horizontal pixel coordinate of the center of a bounding box in a frame, and $l_{foc}$ is the focal length of the RGB camera in pixels \cite{penn-cam-projection}. $l_{foc}$ is not directly specified by CARLA, but can be computed using the image length, 800 pixels, and the camera field of vision, 90 degrees \cite{edmund-focal-length}.

To compute $v^1$ and $v^2$ for detections of the current time step, we also consider detections from the previous time step. First, we attempt to match each bounding box from the current time step to a single bounding box from the previous step. Box pairs are evaluated based on their Intersection-Over-Union (IoU) \cite{simple-real-time-tracking}. Valued between 0 and 1, a high IoU indicates high similarity of size and position of two boxes, and we enforce a minimum threshold of 0.4 for any two boxes to be paired. For two adjacent time steps, A and B, we take the IoU of all possible pairs of bounding boxes with one box from step A, and one from B. These IoU values form the cost matrix for the Hungarian matching algorithm \cite{real-time-mot}, which produces the best possible pairings of bounding boxes from the current time step to the previous.

This matching process results in a set of detections with paired bounding boxes, and a set with unpaired boxes. For each detection with a paired box, we calculate its velocity simply as the difference between respective $d^1$ and $d^2$ values of the current detection and its paired observation from the previous time step, multiplied by the frame rate, $fps_{cam}$. For a detection, $a$, paired with a previous detection, $b$:
\begin{equation}
    <v^1_a, v^2_a> = <d^1_b - d^1_a, d^2_b - d^2_a> * fps_{cam}
\end{equation}

For each detection left unpaired after Hungarian matching, we make no conclusions about $v^1$ or $v^2$ for that detection, and treat each as zero.

Each RADAR measurement output by CARLA represents an additional object detection. RADAR measurements contain altitude ($al$) and azimuth ($az$) angle measurements, as well as depth ($d$) and velocity ($v$), all relative to the RADAR sensor. We convert these measurements into object detection parameters as follows

\begin{align*}
    d^1 &= d * \cos az * \cos al & v^1 &= v * \cos az * \cos al \\
    d^2 &= d * \sin az * \cos al & v^2 &= v * \sin az * \cos al\\
\end{align*}
%end depth/velocity methodology explanation

\section{Derivation of Surrogate Constraints} \label{appendix: surrogate_constraint}

The original attack optimization~\eqref{attack_fake_sp2:obj}-\eqref{attack_fake_sp2:stealthy} may not be convex due to that~\eqref{attack_fake_sp2:target} and~\eqref{attack_fake_sp2:stealthy} could be nonlinear. Our goal in this section is to derive convex surrogate constraints that are good approximations to~\eqref{attack_fake_sp2:target} and~\eqref{attack_fake_sp2:stealthy}. Furthermore, we require the surrogate constraints to be tighter than the original constraints, so that solving the attack under the surrogate constraints will always give us a feasible solution to the original attack. Concretely, we want to obtain surrogate constraints to $F(x)=\ell$, where $\ell\in\{\text{green, yellow, red}\}$. We analyze each case of $\ell$ separately:
\begin{itemize}
\item $\ell=\text{green}$

In this case, $F(x)=\ell$ is equivalent to $v\ge0$ according to~\eqref{eq:FCW}. While this constraint is convex, when we actually solve the optimization, it might be violated due to numerical inaccuracy. To avoid such numerical issues, we tighten it by adding a margin parameter $\epsilon>0$, and the derived surrogate constraint is $v\ge\epsilon$.

\item $\ell=\text{red}$

In this case, $F(x)=\ell$ is equivalent to
\begin{eqnarray}
& v&< 0\label{eq:appendix_red_1}\\
& d&\le -1.2 v+\frac{1}{0.8g}v^2\label{eq:appendix_red_2}.
\end{eqnarray}
Similar to case 1, we tighten the first constraint as 
\begin{equation}
v\le -\epsilon.
\end{equation}
Note that by the first constraint, we must have $v<0$. The second constraint is $d\le -1.2v+\frac{1}{0.8g}v^2$. Given $v<0$, this is equivalent to
\begin{equation}
v\le 0.48g - \sqrt{(0.48g)^2+0.8g d}.
\end{equation}
We next define the following function
\begin{equation}
U(d)=0.48g - \sqrt{(0.48g)^2+0.8g d}.
\end{equation}
The first derivative is
\begin{equation}
U^\prime(d)=-\frac{0.4g}{\sqrt{(0.48g)^2+0.8g d}},
\end{equation}
which is increasing when $d\ge 0$. Therefore, the function $U(d)$ is convex.
We now fit a linear function that lower bounds $U(d)$. Specifically, since $U(d)$ is convex, for any $d_0\ge 0$, we have
\begin{equation}
U(d)\ge U^\prime(d_0)(d-d_0)+U(d_0).
\end{equation}
Therefore, $v\le U^\prime(d_0)(d-d_0)+U(d_0)$ is a tighter constraint than $v\le U(d)$. The two constraints are equivalent at $d=d_0$. Again, we need to add a margin parameter to avoid constraint violation due to numerical inaccuracy. With this in mind, the surrogate constraint becomes 
\begin{equation}\label{eq:appendix_cons_surrogate_red}
v\le U^\prime(d_0)(d-d_0)+U(d_0)-\epsilon,
\end{equation}
Or, equivalently:
\begin{eqnarray}
&v&\le -\epsilon,\\
&v&\le U^\prime(d_0)(d-d_0)+U(d_0)-\epsilon,
\end{eqnarray}
This concludes the proof of our Proposition~\ref{prop:red_surrogate}.
\begin{figure}
\centerline{\includegraphics[width=.35\textwidth]{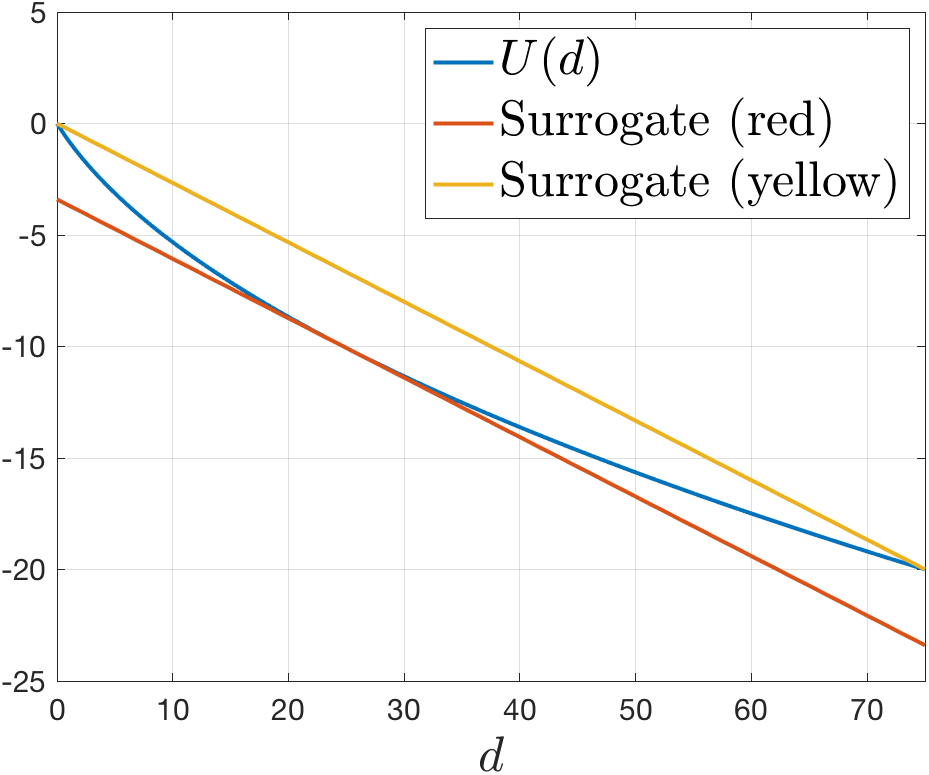}}
\caption{Surrogate light constraints.} 
\label{fig:red_surrogate}
\end{figure}

However, we still need to pick an appropriate $d_0$. In our scenario, the distance $d$ has physical limitation $d\in[\underline d, \bar d]$ with $\underline d=0$ and $\bar d=75$. The  $U(d)$ curve for $d\in [0,75]$  is shown in Fig~\ref{fig:red_surrogate}. Based on the figure, we select $d_0$ such that $U^\prime(d_0)$ is equal to the slope of the segment connecting the two end points of the curve, i.e.,
\begin{equation}
U^\prime(d_0)=\frac{U(75)-U(0)}{75}=\frac{U(75)}{75}.
\end{equation}
We now derive the concrete surrogate constraints used in our experiment section. We begin with the following equation:
\begin{equation}
0.48g+\frac{0.4g}{U^\prime(d)}=U(d).
\end{equation}
From which, we can derive $d_0$:
\begin{equation}
d_0=\frac{1}{0.8g}\left((\frac{30g}{U(75)})^2-(0.48g)^2\right)
\end{equation}
and
\begin{equation}
U(d_0)=0.48g+\frac{30g}{U(75)}.
\end{equation}
By substituting $d_0$ and $U(d_0)$ into~\eqref{eq:appendix_cons_surrogate_red}, we find that the surrogate constraint is
\begin{equation}
v\le \frac{U(75)}{75}(d-d_0)+0.48g+\frac{30g}{U(75)}+\epsilon.
\end{equation}

\item $\ell=\text{yellow}$

In this case, $F(x)=\ell$ is equivalent to
\begin{eqnarray}
& v&< 0\\
& d&\ge -1.2 v+\frac{1}{0.8g}v^2.
\end{eqnarray}
Similarly, we tighten the first constraint to 
\begin{equation}
v\le -\epsilon.
\end{equation}
For the second constraint, the situation is similar to $\ell=\text{red}$. $\forall d_0>0$. We have
\begin{equation}
v\ge \frac{U(d_0)}{d_0}d, \forall d\in [0, d_0]
\end{equation}
The above inequality is derived by fitting a linear function that is always above the $U(d)$ curve. Next, we select $d_0=75$ and add a margin parameter $\epsilon$ to derive the surrogate constraint:
\begin{eqnarray}
& v&\le -\epsilon\\
& v&\ge \frac{U(75)}{75}d+\epsilon.
\end{eqnarray}
To summarize, we have derived surrogate constraints for $F(x)=\ell$, where $l\in \{\text{green, yellow, red}\}$. When we solve the attack optimization, we replace each individual constraint of~\eqref{attack_fake_sp2:target} and~\eqref{attack_fake_sp2:stealthy} by one of the above three surrogate constraints. In Fig~\ref{fig:red_surrogate}, we show the surrogate constraints for red and yellow lights with $\epsilon=10^{-3}$.

\end{itemize}

\section{Preprocessing of CARLA Measurements} \label{appendix: preprocess}
In this section, we describe how we preprocess the measurements obtained from CARLA simulation. The measurement in each time step takes the form of $t_t=[y_t^1, y_t^2]\in \{\R\cup \text{NaN}\}^8$, where $y_t^1\in \{\R\cup \text{NaN}\}^4$ is the vision detection produced by ML-based objection detection algorithm YOLOv2, and $y_t^2$ is the detection generated by radar (details in Appendix \ref{appendix:carla-output-processing}). Both vision and radar measurements contain four components: (1) the distance to MIO along driving direction, (2) the velocity of MIO along driving direction, (3) the distance to MIO along lateral direction, and (4) the velocity of MIO along lateral direction. The radar measurements are relatively accurate, and do not have missing data or outliers. However, there are missing data (NaN) and outliers in vision measurements. The missing data problem arises because the MIO sometimes cannot be detected, e.g., in the beginning of the video sequence when the MIO is out of the detection range of the camera. Outliers occur because YOLOv2 may not generate an accurate bounding box of the MIO, causing it to correspond to a depth map reading of an object at a different physical location. As such, a small inaccuracy in the location of the bounding box could lead to dramatic change to the reported distance and velocity of the MIO. 

In our experiment, we preprocess detections output from CARLA to address missing data and outlier issues. First, we identify the outliers by the Matlab ``filloutliers" method, where we choose ``movmedian" as the detector and use linear interpolation to replace the outliers. The concrete Matlab command is:
$$
\text{\small filloutliers($Y$, `linear', `movmedian', 0, `ThresholdFactor', 0.5)},
$$
where $Y\in\R^{T\times 8}$ is the matrix of measurements and $T$ is the total number steps. In our experiment $T=295$. We perform the above outlier detection and replace operation twice to smooth the measurements. 

\begin{figure}
\centering
	 \begin{subfigure}{.23\textwidth}
		\centering
		\includegraphics[width=1\textwidth, height=0.85\textwidth]{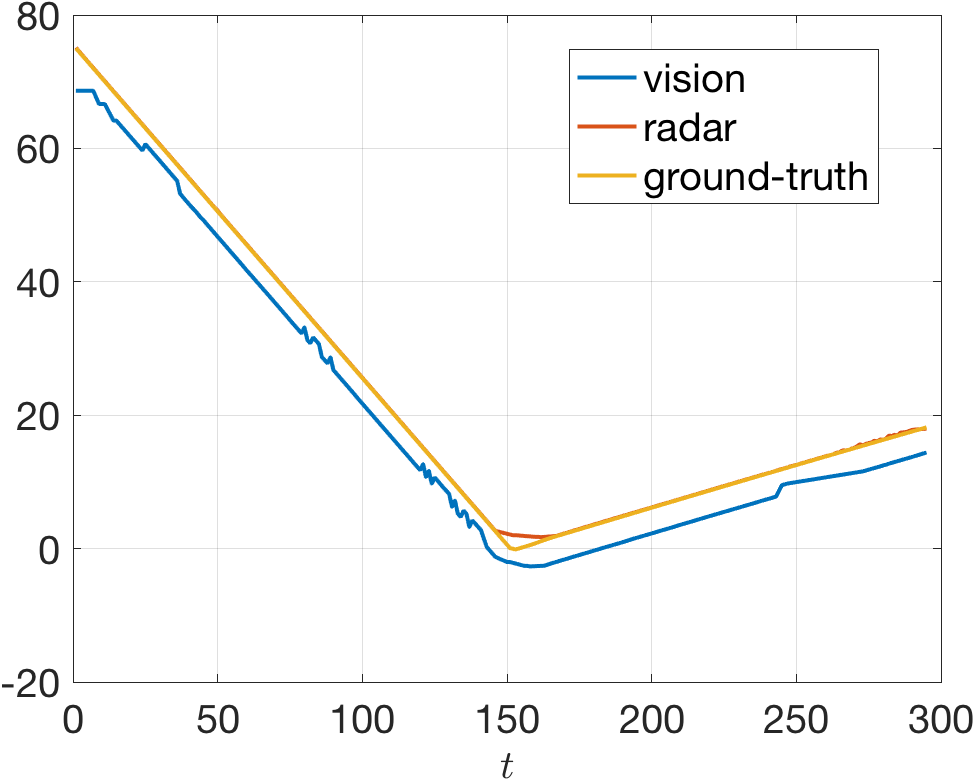}
		\caption{Distance measurements.}
		\label{fig:MIO-10_raw_distance}
	\end{subfigure}
	\hfill
	\begin{subfigure}{.23\textwidth}
		\centering
		\includegraphics[width=1\textwidth, height=0.85\textwidth]{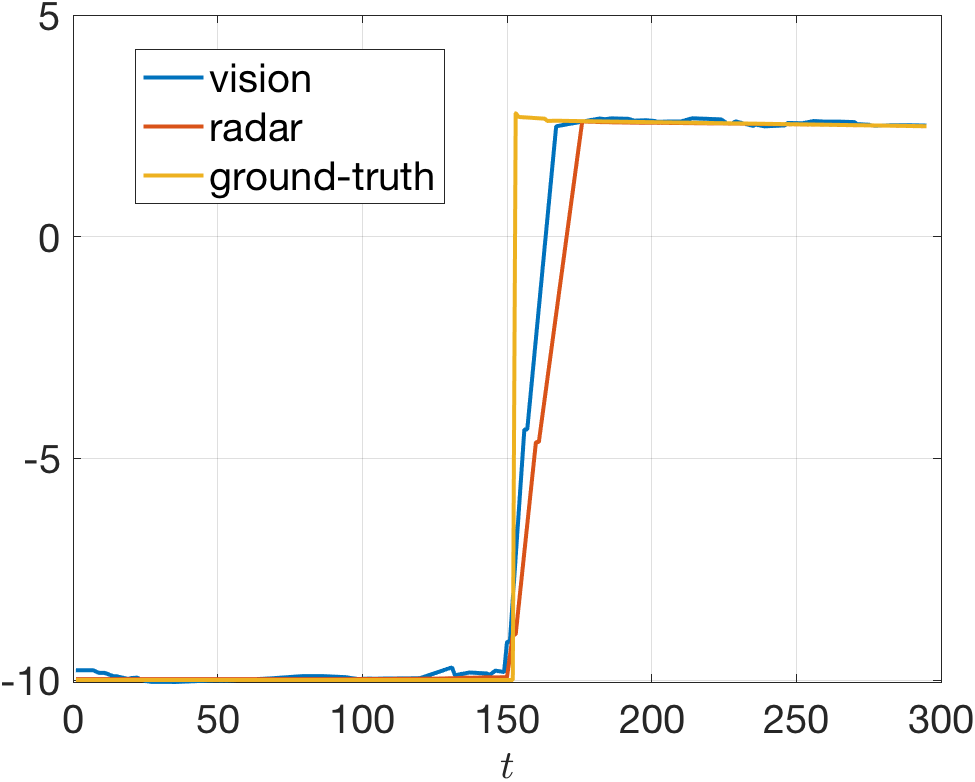}
		\caption{Velocity measurements.}
		\label{fig:MIO-10_raw_velocity}
	\end{subfigure}
	\caption{On the MIO-10 dataset, the preprocessed vision measurements and the radar measurements match the ground-truth reasonably well.}
	\label{fig:MIO-10_raw_measurement}
\end{figure}

\begin{figure}
\centering
	 \begin{subfigure}{.23\textwidth}
		\centering
		\includegraphics[width=1\textwidth, height=0.85\textwidth]{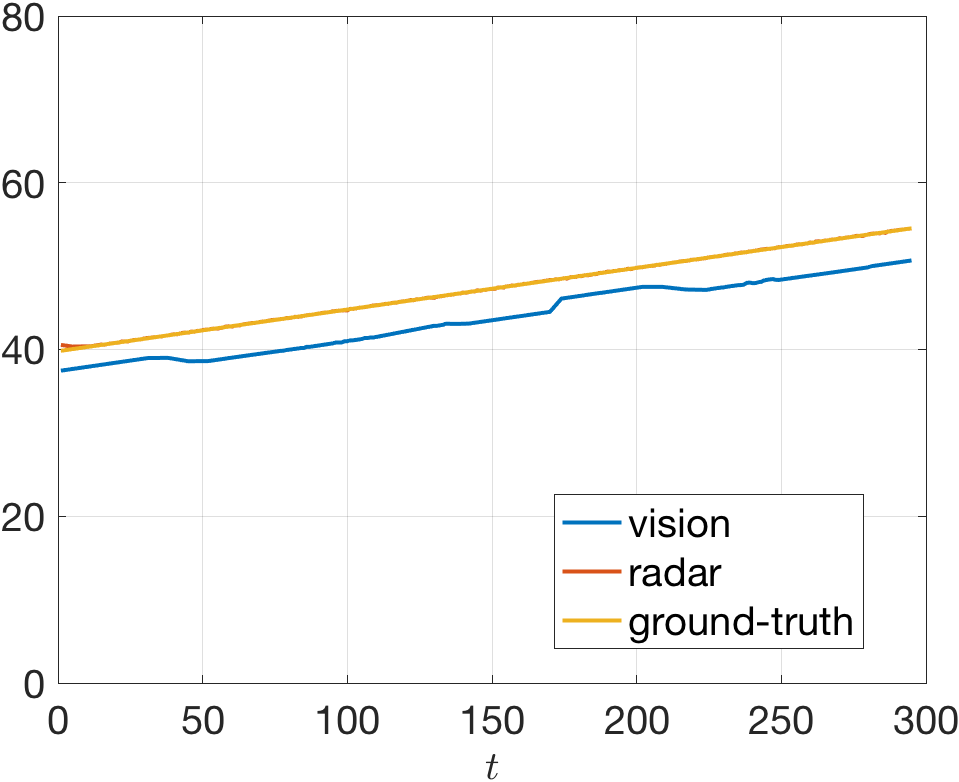}
		\caption{Distance measurements.}
		\label{fig:MIO+1_raw_distance}
	\end{subfigure}
	\hfill
	\begin{subfigure}{.23\textwidth}
		\centering
		\includegraphics[width=1\textwidth, height=0.85\textwidth]{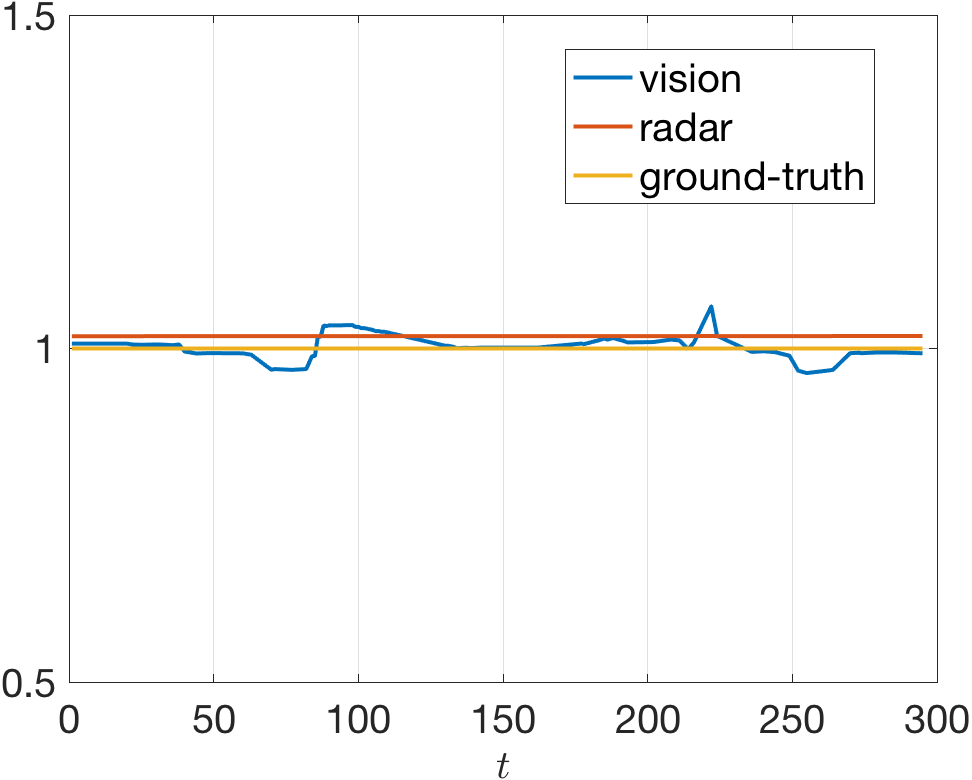}
		\caption{Velocity measurements.}
		\label{fig:MIO+1_raw_velocity}
	\end{subfigure}
	\caption{On the MIO+1 dataset, the preprocessed vision measurements and the radar measurements match the ground-truth reasonably well.}
	\label{fig:MIO+1_raw_measurement}
\end{figure}

Then, we apply the Matlab ``impute" function to interpolate the missing vision measurements. In Fig~\ref{fig:MIO-10_raw_measurement} and~\ref{fig:MIO+1_raw_measurement}, we show the preprocessed distance and velocity measurements from vision and radar compared with the ground-truth for both MIO-10 and MIO+1 datasets. Note that after preprocessing, both radar and vision measurements match with the ground-truth well.

\section{Velocity Increase in Figure~\ref{fig:MIO+1_attack_velocity}} \label{appendix:increase}
\begin{figure}[t]
\begin{subfigure}{.23\textwidth}
		\centering
		\includegraphics[width=1\textwidth, height=0.6\textwidth]{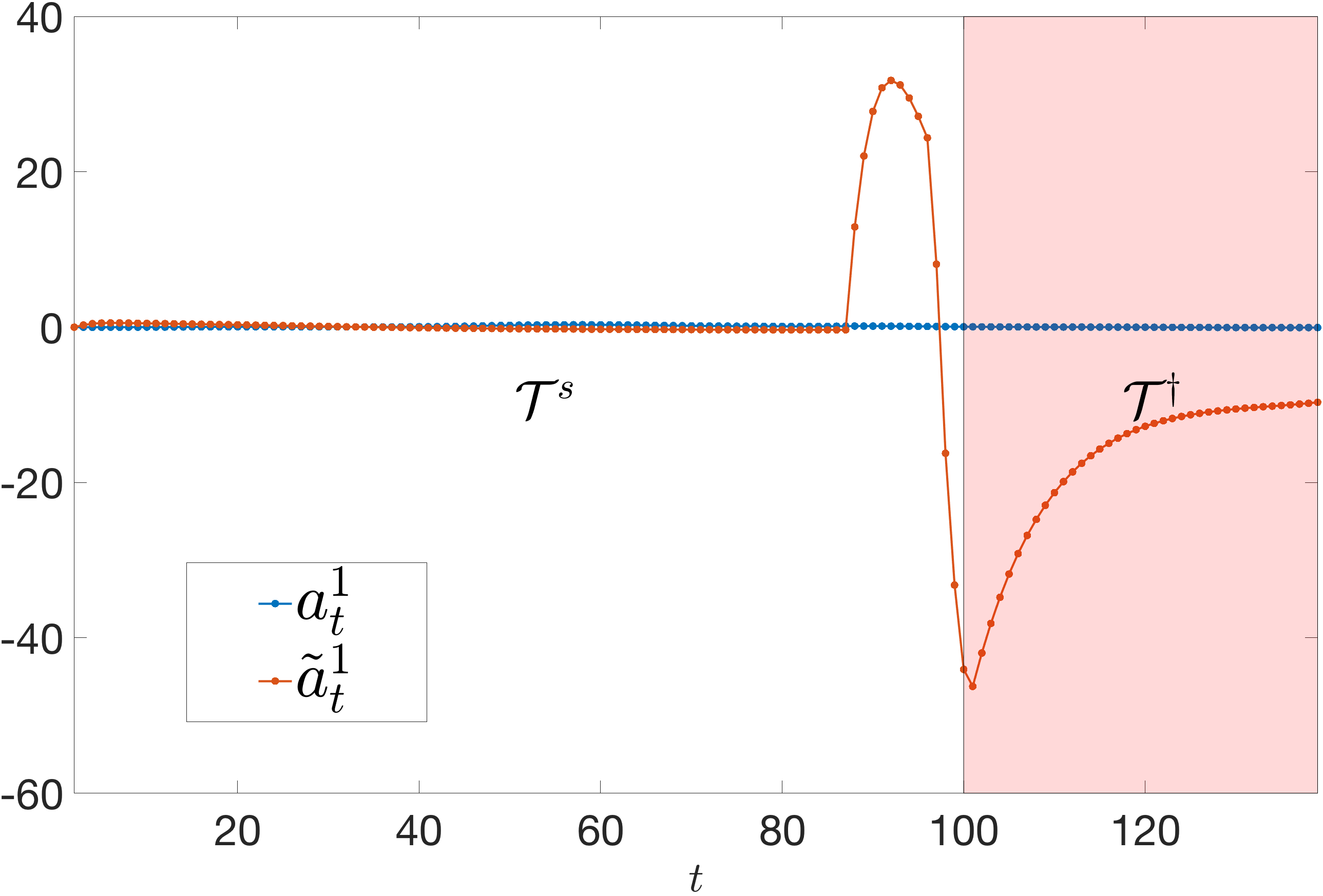}
		\caption{Acceleration estimation.}
		\label{fig:MIO+1_acc}
\end{subfigure}%
\hfill
       \begin{subfigure}{.23\textwidth}
		\centering
		\includegraphics[width=1\textwidth, height=0.6\textwidth]{figures/velocity_measurement_MIO+1.png}
		\caption{Manipulation on velocity.}
		\label{fig:attack_MIO+1_appendix}
	\end{subfigure}%
	\caption{Acceleration reduces significantly as the velocity measurement drops after step 96. This in turn causes the KF velocity estimation to decrease fast.}
\end{figure}

In Figure~\ref{fig:attack_MIO+1_appendix}, we show again the manipulation on the velocity measurement for the MIO+1 dataset. The attacker's goal is to cause the FCW to output red warnings in the target interval $[100,139]$. Intuitively, the attacker should decrease the distance and velocity. However, in Figure~\ref{fig:attack_MIO+1_appendix}, the attacker instead chooses to increase the velocity during interval [88,96]. We note that this is because the attacker hopes to force a very negative KF acceleration estimation. To accomplish that, the attacker first strategically increases the velocity from step 88 to 96, and then starting from step 97, the attacker suddenly decreases the velocity dramatically. This misleads the KF to believe that the MIO has a very negative acceleration. In Fig~\ref{fig:MIO+1_acc}, we show the acceleration estimation produced by KF. At step 96, the estimated acceleration is 8.1m/s$^2$. However, at step 97, the estimated acceleration suddenly drops to $-$16m/s$^2$, and then stays near $-$30m/s$^2$ until the target interval. The very negative acceleration in turn causes the KF velocity estimation to decrease quickly. The resulting velocity estimation reached around $-$10m/s during the target interval, which causes FCW to output red lights.

\section{Human Behavior Algorithm} \label{appendix: human}
In this section, we provide an algorithmic description of the human behavior model.
\begin{algorithm}
\SetAlgoLined
\SetKwInOut{Input}{Input}
\Input{light sequence $\ell_t(1\le t\le T)$, reaction time $h^*$.}
Initialize $s=0$\;
\For{$t\leftarrow 1$ \KwTo $T$}{
\uIf{$t!=1$ and $\ell_{t}!=\ell_{t-1}$}
{
    $s=0$\;
}
\uElseIf{$\ell_t=\text{red}$}
{
    $s=s+1$\;
}
\Else
{
    $s=s-1$\;
}
 \uIf{$s\ge h^*$}
    {
        human applies pressure on pedal\;
    }
   \uElseIf{$s\le -h^*$}
    {
        human releases brake\; 
     }
   \Else{human stays in the previous state\;}
}
 \caption{Human Behavior Algorithm.}
\end{algorithm}

\section{Detailed Results of Greedy Attack} \label{appendix: greedy}
\begin{figure*}
\centering
	 \begin{subfigure}{.24\textwidth}
		\centering
		\includegraphics[width=1\textwidth, height=0.7\textwidth]{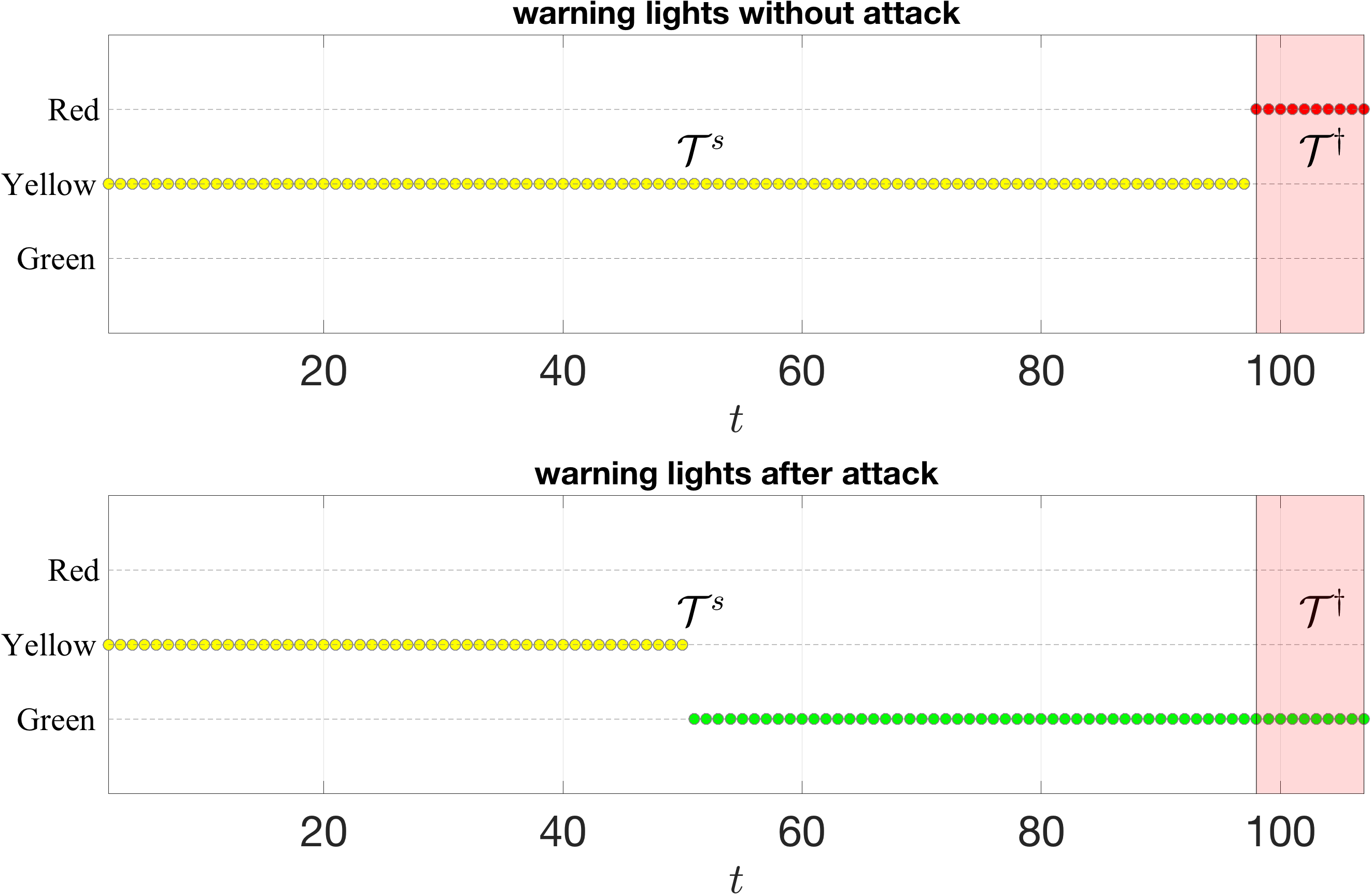}
		\caption{The warning lights.}
		\label{fig:MIO-10_warning_greedy}
	\end{subfigure}%
	\hfill
	\begin{subfigure}{.24\textwidth}
		\centering
		\includegraphics[width=1\textwidth, height=0.7\textwidth]{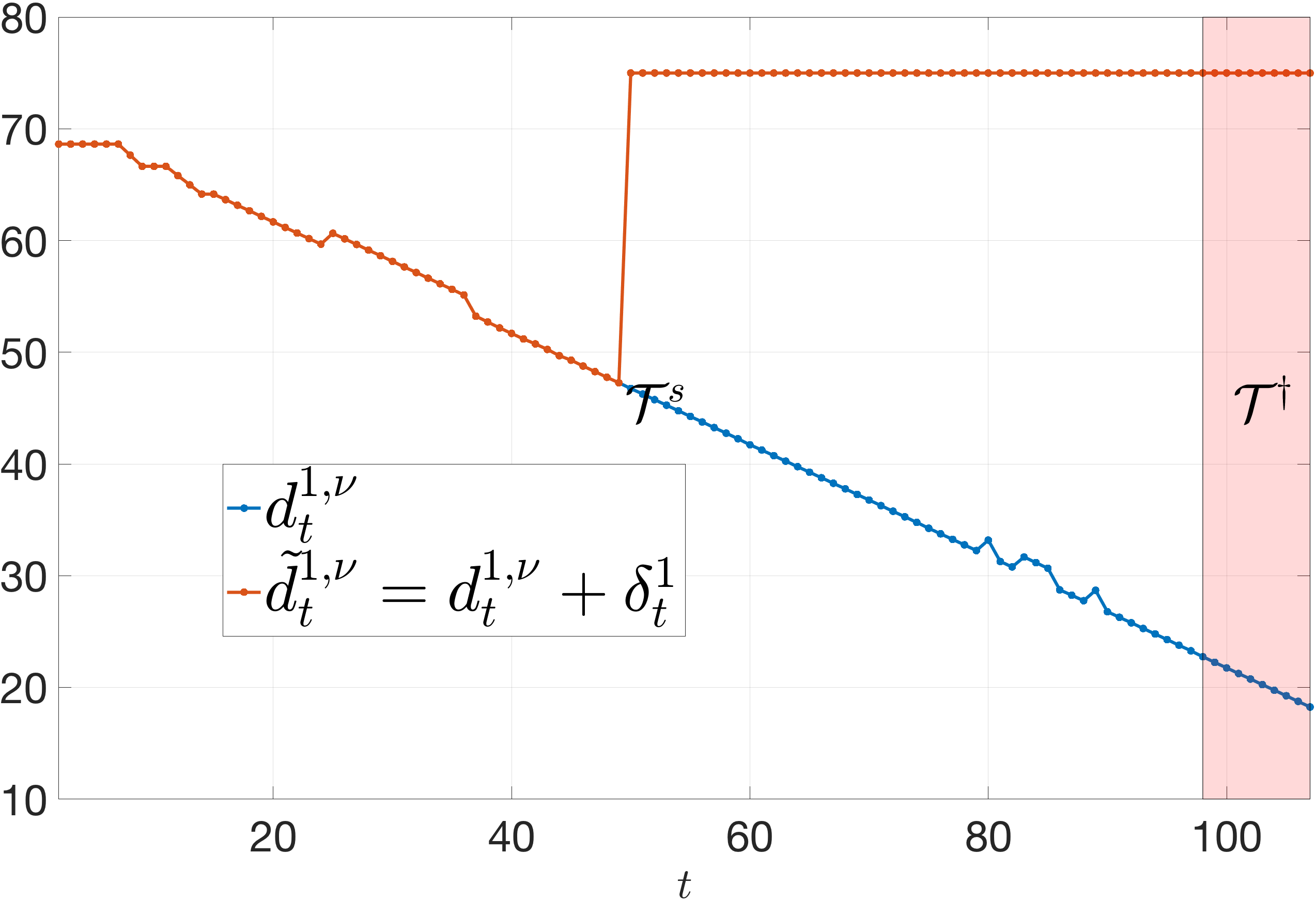}
		\caption{The manipulation on distance.}
		\label{fig:MIO-10_attack_distance_greedy}
	\end{subfigure}%
	\hfill
	\begin{subfigure}{.24\textwidth}
		\centering
		\includegraphics[width=1\textwidth, height=0.7\textwidth]{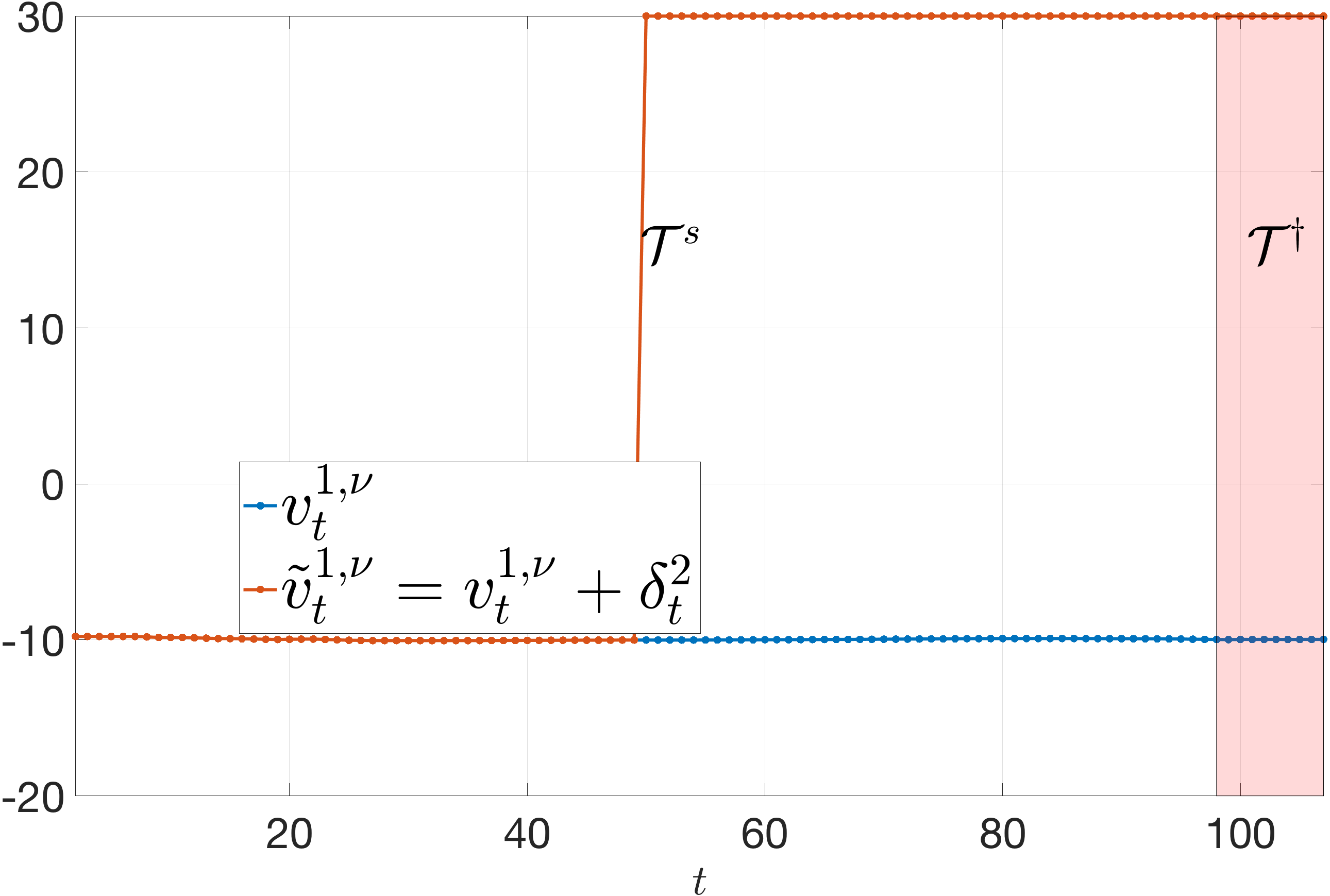}
		\caption{The manipulation on velocity.}
		\label{fig:MIO-10_attack_velocity_greedy}
	\end{subfigure}%
	\hfill
	\begin{subfigure}{.24\textwidth}
		\centering
		\includegraphics[width=1\textwidth, height=0.7\textwidth]{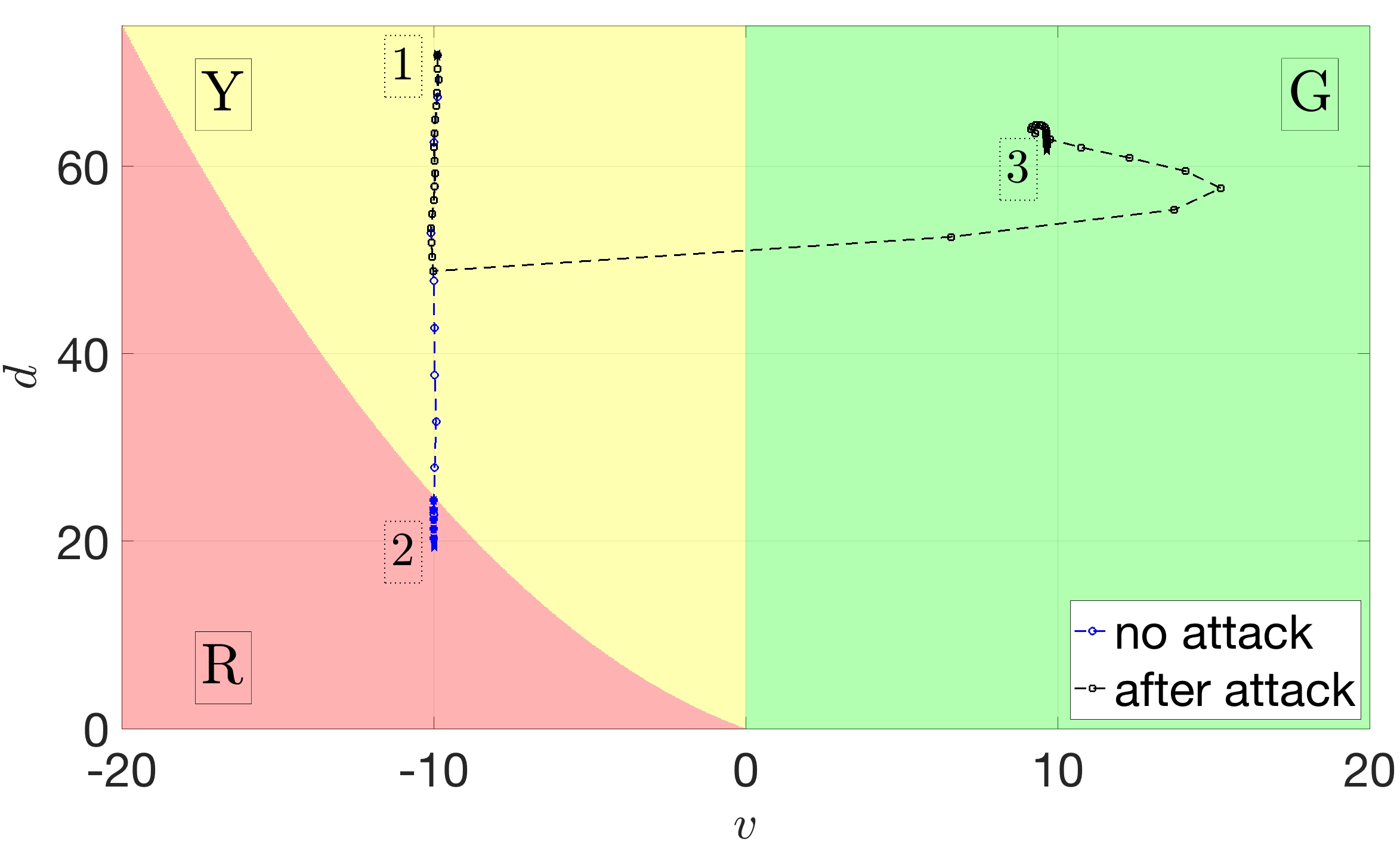}
		\caption{The state trajectory.}
		\label{fig:MIO-10_state_traj_greedy}
	\end{subfigure}%
	\caption{Greedy attack on the MIO-10 dataset.}
	\label{fig:attack_MIO-10_greedy}
\end{figure*}
\begin{figure*}[t]
	 \begin{subfigure}{.24\textwidth}
		\centering
		\includegraphics[width=1\textwidth, height=0.7\textwidth]{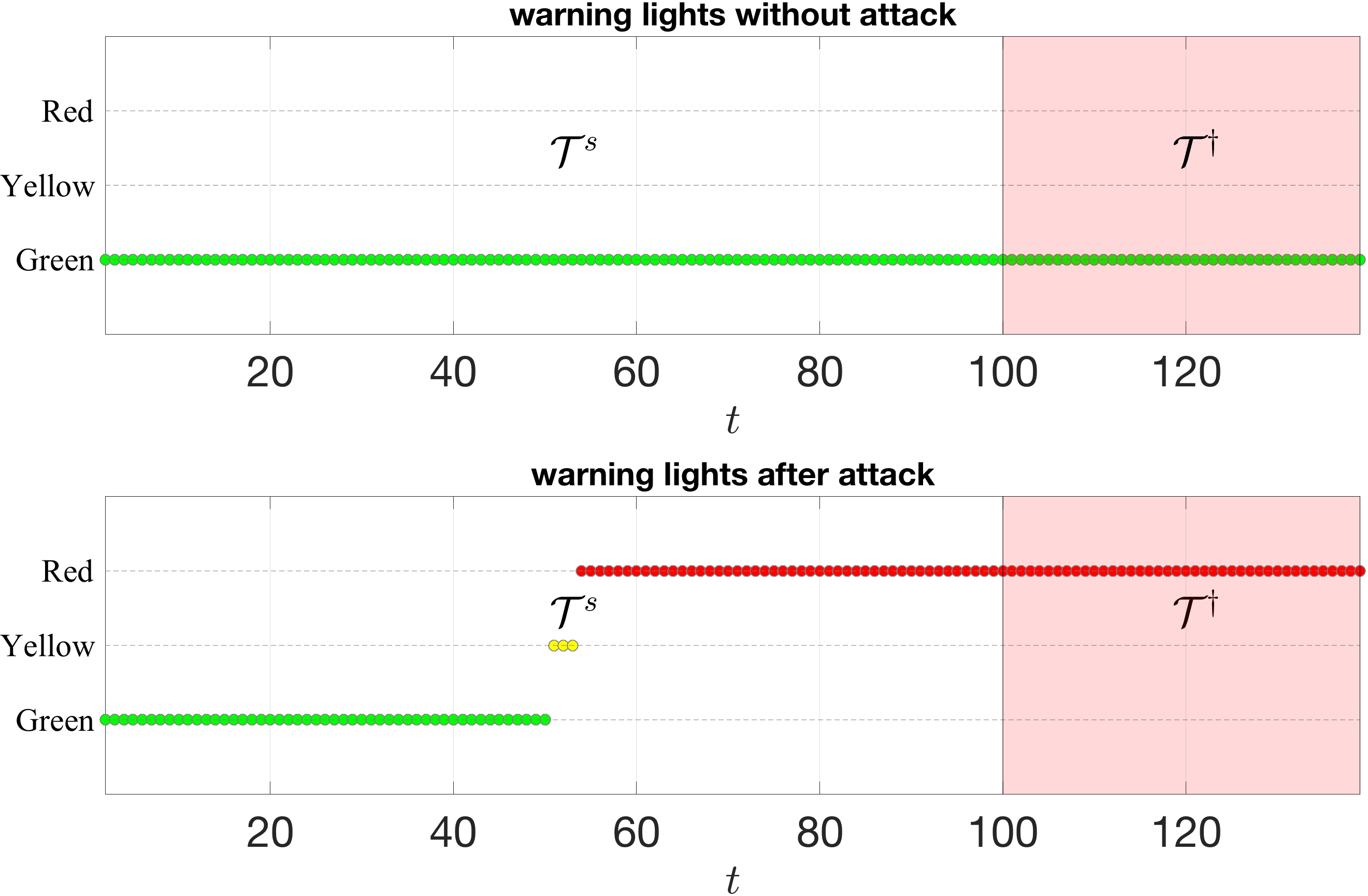}
		\caption{The warning lights.}
		\label{fig:MIO+1_warning_greedy}
	\end{subfigure}%
	\hfill
	\begin{subfigure}{.24\textwidth}
		\centering
		\includegraphics[width=1\textwidth, height=0.7\textwidth]{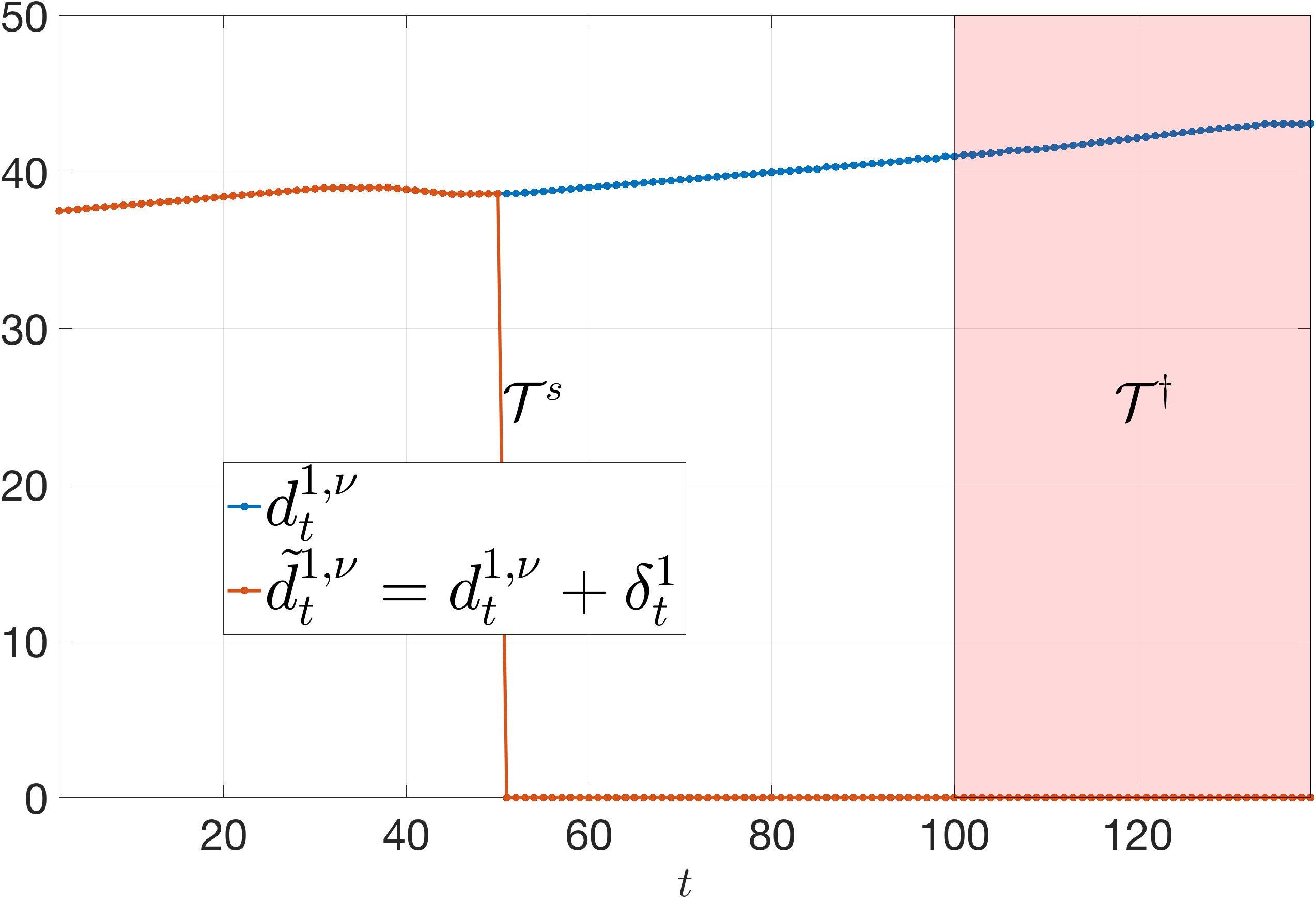}
		\caption{The manipulation on distance.}
		\label{fig:MIO+1_attack_distance_greedy}
	\end{subfigure}%
	\hfill
	 \begin{subfigure}{.24\textwidth}
		\centering
		\includegraphics[width=1\textwidth, height=0.7\textwidth]{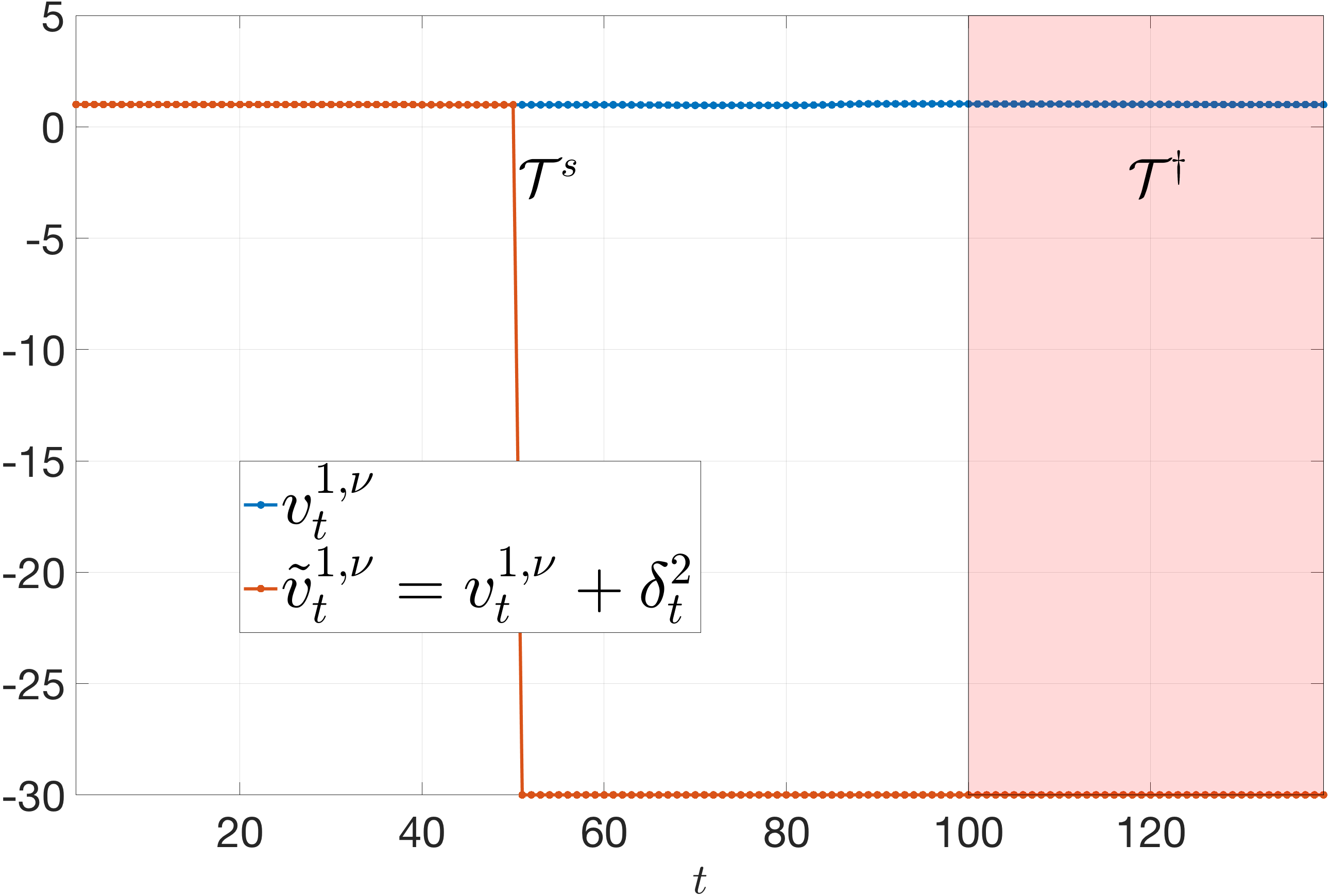}
		\caption{The manipulation on velocity.}
		\label{fig:MIO+1_attack_velocity_greedy}
	\end{subfigure}%
	\hfill
	\begin{subfigure}{.24\textwidth}
		\centering
		\includegraphics[width=1\textwidth, height=0.7\textwidth]{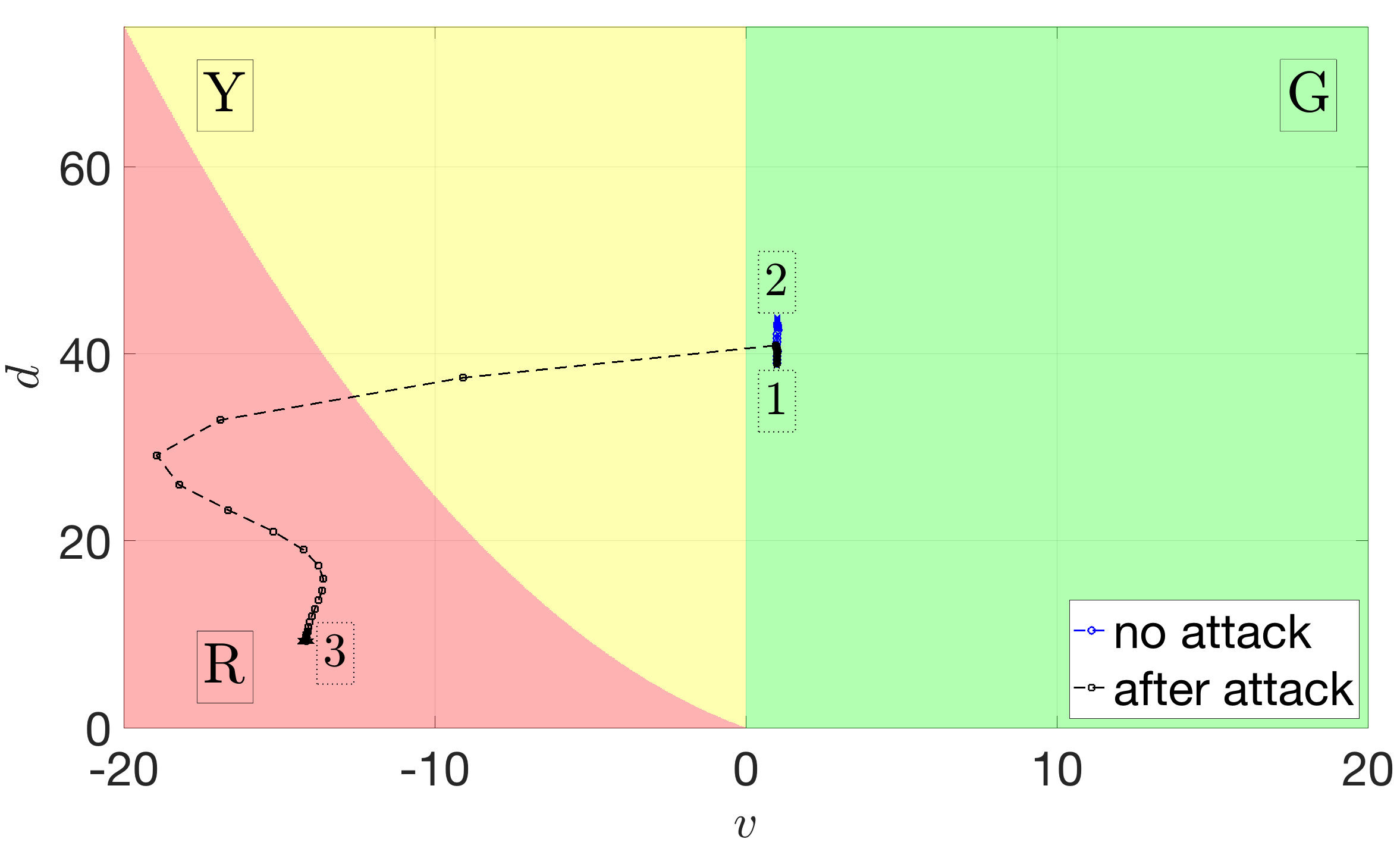}
		\caption{The state trajectory.}
		\label{fig:MIO+1_state_traj_greedy}
	\end{subfigure}%
	\caption{Greedy attack on the MIO+1 dataset.}
	\label{fig:attack_MIO+1_greedy}
\end{figure*}

In this section, we provide more detailed results of the greedy attack, including warning lights before and after attack, manipulations on measurements, and the trajectory of KF state predictions. We notice that the results are very similar for different lengths of the stealthy interval $\T^s$. Therefore, here we only show the results for $\T^s=2.5$ seconds (i.e., half of the full length) as an example.

Fig.~\ref{fig:attack_MIO-10_greedy} shows the greedy attack on MIO-10, where the stealthy interval $\T^s=[50, 97]$. By manipulating the distance and velocity to the maximum possible value, the attacker successfully causes the FCW to output green lights in the target interval $\T^\dagger$. However, the attack induces side effect in $\T^s$, where the original yellow lights are changed to green. In contrast, our MPC-based attack does not have any side effect during $\T^s$. Also note that the trajectory of the KF state prediction enters ``into" the desired green region during $\T^\dagger$. This is more than necessary and requires larger total manipulation ($J_1$) than forcing states just on the boundary of the desired region, as does our attack.

In Fig.~\ref{fig:attack_MIO+1_greedy}, we show the greedy attack on MIO+1. The stealthy interval is $\T^s=[51, 99]$. Again, the attack results in side effect during the stealthy interval $\T^s$. Furthermore, the side effect is much more severe (green to red) than that of our MPC-based attack (green to yellow). The KF state trajectory enters ``into" the desired red region, and requires larger total manipulation ($J_1$) than our attack.

\end{document}